\documentclass[preprint,12pt]{elsarticle}

\usepackage{amssymb}
\usepackage{graphicx}
\usepackage{amsmath}
\usepackage{amssymb}
\usepackage{booktabs}

\usepackage[pagebackref,breaklinks,colorlinks]{hyperref}
\usepackage{times}
\usepackage{epsfig}
\usepackage{graphicx}
\usepackage{amsmath}
\usepackage{amssymb}
\usepackage{adjustbox}
\usepackage{tabularx}
 \usepackage{verbatim}
\usepackage{bm}
\usepackage{subcaption}

 \usepackage{algorithm}

\usepackage{algorithmic}%

\usepackage[capitalize]{cleveref}
\crefname{section}{Sec.}{Secs.}
\Crefname{section}{Section}{Sections}
\Crefname{table}{Table}{Tables}
\crefname{table}{Tab.}{Tabs.}

\journal{Neurocomputing}

\begin{document}

\begin{frontmatter}



\title{Continuous Unsupervised Domain Adaptation Using Stabilized Representations and Experience Replay}


\author{Mohammad Rostami}

\affiliation{organization={University of Southern California}
}
\begin{abstract}
We  introduce an algorithm for tackling the problem of unsupervised domain adaptation (UDA) in continual learning (CL) scenarios. The primary objective is to maintain model generalization under domain shift when new domains   arrive  continually  through updating a base model when only unlabeled data is accessible in subsequent tasks.  While there are many existing UDA algorithms, they typically require access to both the source and target domain datasets simultaneously. Conversely, existing CL approaches can handle tasks that all have labeled data. Our solution is based on stabilizing the learned internal distribution to enhances the model generalization on new domains. The internal distribution is modeled by network responses in hidden layer. We model this internal distribution using a Gaussian mixture model (GMM ) and update the model by matching the internally learned distribution of new domains to the estimated GMM. Additionally, we leverage experience replay to overcome the problem of catastrophic forgetting, where the model loses previously acquired knowledge when learning new tasks. We offer theoretical analysis to explain why our algorithm would work. We also offer extensive comparative and analytic experiments to demonstrate that our method is effective. We perform experiments on four benchmark datasets to demonstrate that our approach is effective. Our implemntation code is available at: \url{https://github.com/rostami-m/LDACID}. \footnote{This work is based on results partially presented at the 2021 Conference on Advances in neural information processing systems   \cite{rostami2021lifelong}.}
\end{abstract}






\end{frontmatter}




\section{Introduction}

Deep neural networks relax the need for manual feature engineering by learning to generate discriminative features in an end-to-end blind training procedure  \cite{morgenstern2014properties,lecun2015deep,liu2018feature}. Despite significant advances in deep learning, however, robust \textit{generalization} of deep neural networks on unseen data is still a primary challenge when \textit{domain shift} exists between the training and the testing data  \cite{tzeng2017adversarial,jia2019domain,wei2021online}. Due to temporal nature of most problems over a long enough period, the distribution of test data undergoes temporal changes, know as \textit{domain shift} in the literature~ \cite{sun2016return,amodei2016concrete}.
Domain shift is a natural challenge in \textit{continual learning} (CL)  \cite{zenke2017continual,chen2018lifelong,lenga2020continual} where the goal is to learn adaptively and autonomously when the underlying input distribution drifts over extended time periods.
Distributional shifts over time usually lead to performance degradation of trained models because of distributional gap between the testing and the training data. To maintain model performance,  retraining becomes necessary to acquire knowledge about new distributions.

Retraining a model from scratch would require persistent manual data annotation which is practically infeasible because of being an economically costly and time-consuming process  \cite{rostami2018crowdsourcing,rasmussen2022challenge}. 
Fine-tuning the trained network may require less training data compared to a network initialized with random weights~ \cite{sharif2014cnn}, but still, data annotation would be necessary. Hence, it cannot be served as a sustainable solution in CL settings.
On the other hand, most CL algorithms  consider tasks, i.e., domains, with fully labeled datasets. Hence, these algorithms still require annotating   training datasets for new observed domains. To relax the need for persistent data annotation under a continuous domain shift, our goal is to develop an algorithm for continual adaptation of a model for tackling the challenge of domain shift in a CL setting using solely unannotated datasets.

Unsupervised Domain adaptation (UDA) is a highly relevant learning setting to our problem of interest. The classic UDA setting involves transferring knowledge from a source domain to  a target domain when both domains share the same classes. The goal  is to train a model for the \textit{target domain} with unannotated data by transferring knowledge from the related \textit{source domain} in which annotated data is accessible  \cite{tzeng2017adversarial,oza2023unsupervised,ge2023unsupervised}. A primary group of UDA algorithms map the training data points for both domains into a shared latent embedding space and align the distributions of the source and the target domains in that space to make the extracted features domain-agnostic  \cite{chen2019joint,dou2019domain,peng2020domain2vec}. Hence, a source-trained classifier that receives its input from the shared embedding space would generalize on the target domain as well. The domain alignment procedure has been implemented either using generative adversarial learning  \cite{he2016deep,sankaranarayanan2018generate,pei2018multi,zhang2019domain,long2018conditional,jian2023unsupervised,hassanpour2023survey} or by directly minimizing the distance between the two distributions  \cite{long2015learning,ganin2014unsupervised,long2017deep,kang2019contrastive,ge2023unsupervised}.

UDA methods that are based on generative adversarial learning  \cite{he2016deep,sankaranarayanan2018generate,pei2018multi,zhang2019domain,long2018conditional}, model the matching deep encoder as a generative network which is trained to extract discriminative features from the two domains that become indistinguishable by a competing discriminative network. This process aligns the two distributions indirectly because the extracted features become relatively identical. Methods that align the distributions directly  \cite{long2015learning,ganin2014unsupervised,long2017deep,kang2019contrastive,rostami2019deep,rostami2020generative}, use a proper loss function that measures the distance between two distributions and minimize the loss function. As a result, the two distributions aligned directly when their distances is minimized. Both approaches for UDA have been used successfully. In our work, we rely on the direct domain alignment approach based on metric minimization.

The second approach for domain alignment in UDA is to use a probability distance to measure distributional discrepancy and then train an encoder to minimize the cross-domain distance at its output as the latent shared embedding space~ \cite{ghifary2016deep,morerio2017minimal,damodaran2018deepjdot,lee2019sliced,stan2021unsupervised,rostami2019deep}. 
 The key question is to select the right metric for this purpose. The Maximum Mean Discrepancy (MMD) is one of the early metrics has been employed to align the means of two distributions for probability matching  \cite{long2015learning,long2017deep}. Building upon this baseline, Sun et al.  \cite{sun2016deep} introduced an improvement by aligning distribution correlations, thereby incorporating second-order statistics. Other advancements include the use of the central moment discrepancy  \cite{zellinger2016central} and adaptive batch normalization  \cite{li2018adaptive}.
 
While domain alignment based on aligning the lower order probability moments is straightforward, it overlooks mismatches in higher moments. To address this limitation, the Wasserstein distance (WD) has been utilized to capture information from higher-order statistics  \cite{courty2017optimal,damodaran2018deepjdot}. Damodaran et al.  \cite{damodaran2018deepjdot} demonstrated that employing WD yields improved Unsupervised Domain Adaptation (UDA) performance compared to using MMD or correlation alignment  \cite{long2015learning,sun2016deep}.
There are also theoretical guarantees that demonstrates that domain alignment using WD will minimize an upperbound for the target domain expected error~ \cite{redko2017theoretical}.
WD metric offers a significant advantage in deep learning in general~ \cite{arjovsky2017wasserstein} because unlike more common probability metrics such as KL-divergence or JS-divergence, the Wasserstein distance possesses non-vanishing gradients even when the two distributions do not have overlapping supports~ \cite{kolouri2016sliced,rostami2019sar,damodaran2018deepjdot}.  Since the two distributions might not have overlapping support when UDA problem is solved, this property makes WD well-suited for deep learning as it can be effectively minimized using first-order optimization methods.
This is particularly advantageous since deep learning objective functions are typically optimized using gradient-based techniques. The downside of using WD is that it does not have a closed form solution and is defined as the solution of an optimization problem. In this work, we   rely on  the sliced Wasserstein distance (SWD)~ \cite{lee2019sliced} variant of the Wasserstein distance  because it can be computed efficiently using a closed form solution. Moreover, this computation can be done using only the empirical samples of the two data distributions. Since in practice, we only have the samples of the two distributions, SWD is suitable choice for UDA.

Most existing UDA algorithms are not suitable for continual learning because the underlying model can be trained if datasets from both domains are accessible at the same time. In other words, the domains are not learned sequentially. There are a few methods that address UDA in a sequentially setting, where it is assumed the source data is not accessible during training time~ \cite{stan2021domain,rostami2023overcoming,kundu2020universal,yang2021generalized,zhang2023source,shenaj2023learning}. However, these methods usually consider only a single target domain and a single source domain and the primary goal is preforming well in the target domain. Finally, simply updating the underlying model to generalize in the currently encountered domain is not sufficient. Because upon updating the model, the network is likely to forget the past learned domains as a result of retroactive interference, referred to as the phenomenon of \textit{catastrophic forgetting}  \cite{french1999catastrophic,kirkpatrick2017overcoming} in the continual learning literature. Consequently, we also need to tackle catastrophic forgetting in continual domain adaptation setting.

 Existing CL methods primarily employ two main approaches to tackle catastrophic forgetting: model regularization and experience replay techniques  \cite{kirkpatrick2017overcoming, zenke2017continual}.
Model regularization techniques aim to identify network weights that are crucial for retaining knowledge about previously learned tasks. These important weights are then consolidated during model updates to ensure that they contribute to learning subsequent tasks while preserving the knowledge of past tasks  \cite{kirkpatrick2017overcoming, zenke2017continual, aljundi2018memory,srinivasan2022climb}. By consolidating the important weights, the model can better adapt to new tasks without significantly forgetting previously acquired knowledge. However, a drawback of model regularization is that as more tasks are learned, the learning capacity of the network may become constrained, as more and more weights are consolidated.
An alternative approach is progressive expansion, where the model is gradually expanded by adding new weights to accommodate the learning of new tasks  \cite{rusu2016progressive, xu2021adaptive,cai2023task}. These new weights are added such the past weights can help knowledge transfer for learning the current task. Both model expansion and model regularization  strategies allow for the separation of information pathways for distinct tasks, minimizing interference between different task representations and reducing the likelihood of catastrophic forgetting.
 
 Drawing inspiration from existing works in UDA and CL, we propose an algorithm that enables continual unsupervised adaptation of a model to new domains using only unannotated data.  Our method can be considered as an improvement over existing UDA and CL methods, where challenges from both learning setting are addressed.
 We would like to highlight that while domain adaptation under continuous domain shift is not studied significantly in the literature, there are a few relevant works. Wulfmeier et al.  \cite{wulfmeier2018incremental} investigate the problem of gradual shifts in evolving environments. While their work is relevant, it does not specifically address the issue of domain shift in continual learning settings.
On the other hand, Bobu et al.  \cite{bobu2018adapting} and Wu et al.  \cite{wu2019ace} delve into the problem of domain shift in continual learning. However, these works make the assumption that datasets for all learned tasks are observable at each time-step. This assumption greatly limits the practicality of their approaches, as it is often infeasible to have access to all previously learned tasks' datasets simultaneously.
Another related work by Porav et al.  \cite{porav2019don} explores a similar setting to ours. However, their proposed method relies on image translation techniques, which makes the approach highly domain-specific and may not generalize well to different types of data.
In contrast to these existing methods, our work overcomes these limitations and presents a more general algorithm. Our method is not reliant on specific domain assumptions and can effectively adapt to shifting distributions across sequentially learned tasks without requiring access to all previous task datasets at once.

 Our approach is centered around the idea of consolidating the internally learned distribution, which captures the knowledge acquired by the model during the initial learning phase on a  source domain. More specifically, when a model is trained to solve a classification problem, it learns to map the data distribution in the input space to an internal distribution, represented by network responses in its hidden layers.  The internal distribution is multimodal distribution, where each class is represented by a mode~ \cite{rostami2023domain,pan2019transferrable,snell2017prototypical}. The internal distribution can serve as a basis for updating the model in a way that we ensure the learned internal distributions for all subsequent unannotated domains are shared. As a result, the effect of domain shift  will be mitigated internally.

To address the issue of catastrophic forgetting, we  benefit from experience replay~ \cite{schaul2015prioritized,jin2021learn}. The fundamental idea behind experience replay is to identify crucial training data points that significantly contribute to learning each task and store them in a memory buffer  \cite{schaul2015prioritized,mirtaheri2023history}. These stored samples serve as a representative set of the previously learned distributions.
During subsequent training steps, these stored samples are replayed alongside the current task's data. By including these samples during training, the model has the opportunity to revisit and learn from the previously encountered distributions, allowing it to retain the acquired knowledge about past tasks  \cite{robins1995catastrophic, shin2017continual,goodfellow2013empirical}. 
To mitigate the difficulty of sample selection, some CL algorithms employ generative experience replay. Instead of storing actual samples in the memory buffer, these methods generate synthetic pseudo-samples that closely resemble the samples from previously learned tasks  \cite{shin2017continual,kamra2017deep,rostami2019complementary,rostami2020generative,rostami2021cognitively}. This approach provides a solution to the sample selection problem by allowing the model to access representative samples from past tasks through generative synthesis.

In our work, we  store representative samples from each task and replay them along with the current task data when updating the model. This strategy helps alleviating the problem of forgetting previously learned domains, enabling the model to retain the past acquired knowledge while adapting to new domains.
The major challenge in using the experience replay approach is the selection of important data points to include in the memory buffer. To address this challenge, we leverage the consolidated internal distribution that we have developed in our method. 
To support the effectiveness of our approach, we provide a   theoretical analysis demonstrating how our method mitigates catastrophic forgetting and enhances generalization performance on all domains learned so far. Additionally, we evaluate the performance of our method on well-established UDA benchmark datasets, showcasing its efficacy when compared against existing methods.

\section{Problem Statement}
Consider a classification problem in a given source domain $\mathcal{S}$, where we have access to a labeled training dataset $\mathcal{D}_{\mathcal{S}}=(\bm{X}_{0}, \bm{Y}_{0})$. The input data consists of $\bm{X}_{0}=[\bm{x}_1^0,\ldots,\bm{x}_N^0]\in\mathcal{X}\subset\mathbb{R}^{d\times N}$, and the corresponding labels are $\bm{Y}_{0}=[\bm{y}^0_1,\ldots,\bm{y}^0_N]\in \mathcal{Y}\subset\mathbb{R}^{k\times N}$. The training samples are independently drawn from an unknown source distribution, denoted as $\bm{x}_i^0\sim p_{0}(\bm{x})$. To find an optimal model, we utilize a deep neural network $f_\theta$ with learnable weights $\theta$ and apply empirical risk minimization (ERM):
\begin{equation}
\hat{\theta}_0=\arg\min{\theta}\sum_i \mathcal{L}(f_{\theta}(\bm{x}_i^0),\bm{y}i^0) 
\end{equation}
 where $\mathcal{L}(\cdot)$ represents a suitable loss function such as cross entropy.

 When the training dataset is sufficiently large and the network structure is complex, the ERM procedure leads to a model that generalizes well on unseen data points drawn from the training data distribution $p_{0}(\bm{x})$  \cite{shalev2014understanding}.
However, the input distribution is subject to non-stationarity in CL setting, meaning that the testing samples can be drawn from drifted versions of the training distribution. This distributional shift introduces a gap that hampers the model's generalization performance during testing. Our objective is to continually update the source-trained model $f_{\hat{\theta}_0}$ using only unlabeled data to address the issue of poor generalization without forgetting the knowledge acquired from past experiences. To formalize this process, we consider a series of sequentially arriving target domains $\mathcal{T}^t, t= 1\ldots T$, each accompanied by an unlabeled dataset $\mathcal{D}_{\mathcal{T}}^t=(\bm{X}_t)$, where $\bm{X}_t \in\mathbb{R}^{d\times M_t}$ and $\bm{x}_i^t\sim p_{t}(\bm{x})$. Note that the distributions $p_t(\bm{x})$ are distinct for each target domain, i.e., $\forall t_1, t_2: p_{t_1}\neq p_{t_2}$. In this setting, using ERM alone is not feasible since the target domains are unlabeled. However, since we encounter the drifted versions of the training distribution, we can benefit from the cross-domain similarities to update the model to adapt to the drifted versions of the input distribution.

Traditional UDA methods are inadequate for addressing CL settings as they typically require access to all the training datasets to solve a joint optimization problem.  However, this joint-training approach is not practical in a CL setting where data arrives sequentially. Additionally, most UDA methods only consider a single target domain and extending them to continual model adaptation settings in not trivial. Therefore, alternative strategies are needed to tackle the challenges of CL in the context of unsupervised domain adaptation.
 To address challenges of ``domain shift'' and ``catastrophic forgetting'', we consider that the base  network $f_\theta(\cdot)$ can be decomposed into a  deep encoder $\phi_{\bm{v}}(\cdot): \mathcal{X}\rightarrow \mathcal{Z}\subset \mathbb{R}^p$ and a classifier subnetwork $h_{\bm{w}}(\cdot): \mathcal{Z}\rightarrow \mathcal{Y}$, i.e., $f_\theta = h_{\bm{w}}\circ \phi_{\bm{v}}$, where $\theta=(\bm{w},\bm{v})$. Our method is based on consolidating the internal distribution that is formed in the embedding space  $\mathcal{Z}$. We assume that as the result of initial training on the source domain, the  source classes become separable in $\mathcal{Z}$.   If at each time-steps, we update the model  such that the internal distribution remains stable, i.e., the distance between the distributions $\phi(p_0(\bm{x}^0))$ and $\phi(p_t(\bm{x}^t))$ is minimized, then  the model continues to generalize well on the target domains, despite initially  being trained with the source domain labeled dataset. 
This strategy has been used extensively by the existing UDA algorithms but we are constrained with the accessibility of $\mathcal{D}_{\mathcal{S}}$ in CL, i.e., the term $\phi(p_0(\bm{x}^0))$ cannot be  computed. Additionally, we need to update the model such that the new learned knowledge does not interfere with the past learned knowledge.

\section{Proposed Solution}
\label{sec:UDAproposdframework}

\begin{figure}[ht!]
    \centering
    \includegraphics[width=.95\linewidth]{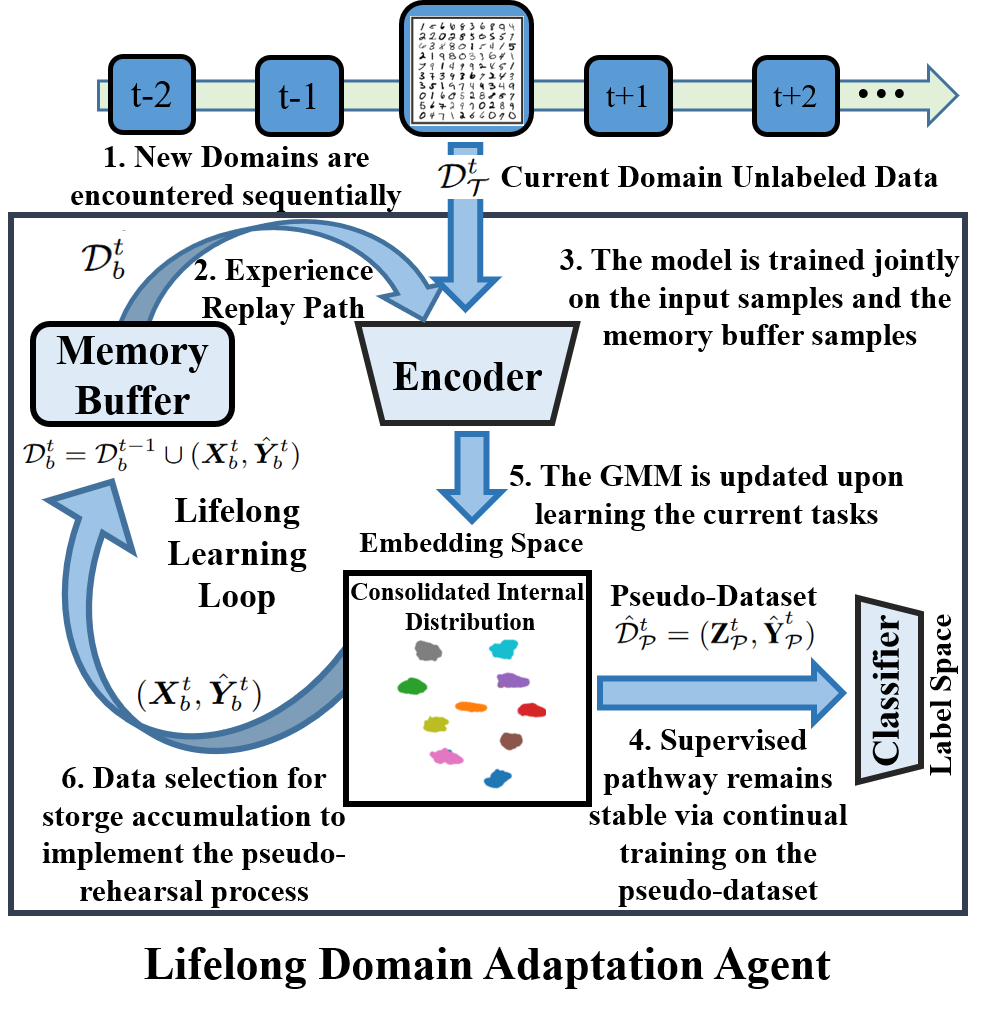}
         \caption{Architecture and learning procedure for the proposed continual UDA framework: the figure captures the steps that are carried out at each time-step $t>0$. At each time-step, six steps are performed sequentially according to the shown order.}
         \label{figDL:LL}
\end{figure}

Our approach is centered around consolidating the intermediate distribution acquired through discriminative embedding, aiming to preserve the generalizability of the model. After learning the initial source domain, the encoder undergoes training to map the input source distribution to a multi-modal distribution $p_J(\bm{z})$ in the embedding space. Each mode within this distribution corresponds to one of the input classes, and the training data points belonging to a particular class are mapped to the same cluster (refer to Figure~\ref{figDL:LL}, middle). The internally learned multi-modal distribution is empirically represented by the data representations of the source domain, denoted as $\{(\phi_{\bm{v}}(\bm{x}_i^{0}),\bm{y}i^{0})\}_{i=1}^{N}$. To estimate this internal distribution, we utilize a parametric model called Gaussian Mixture Model (GMM) denoted as $p_J^0(\bm{z})$, which consists of $k$ components:
\begin{equation}
\footnotesize
p_J^0(\bm{z})=\sum_{j=1}^k \alpha_j^0
\mathcal{N}(\bm{z}|\bm{\mu}_j^0,\bm{\Sigma}_j^0),
\end{equation}  
where the mixture weights are denoted by $\alpha_j^0$, while $\bm{\mu}_j^0$ and $\bm{\Sigma}_j^0$ represent the mean and covariance matrices for each component, respectively. The computation of these parameters typically involves an iterative procedure using the expected maximization (EM) algorithm. However, in our case, since we have access to the labels, we can compute the parameters for each mode independently. Let $\bm{S}_j^0$ denote the support set for mode $j$, which can be defined as $\bm{S}_j^0=\{(\bm{x}_i^0,\bm{y}i^0)\in \mathcal{D}_{\mathcal{S}}|\arg\max\bm{y}_i^0=j \}$. We can utilize a maximum a posteriori (MAP) estimation, we can obtain the following estimates for the GMM parameters: 
\begin{equation}
\footnotesize
\begin{split}
&\hat{\alpha}_j^0 = \frac{|\bm{S}_j^0|}{N}\\&\hat{\bm{\mu}}_j = \sum_{(\bm{x}_i^0,\bm{y}_i^0)\in \bm{S}_j^0}\frac{1}{|\bm{S}_j^0|}\phi_v(\bm{x}_i^0), \\&\hat{\bm{\Sigma}}_j^0 =\sum_{(\bm{x}_i^0,\bm{y}_i^0)\in \bm{S}_j^0}\frac{1}{|\bm{S}_j^0|}\big(\phi_v(\bm{x}_i^0)-\hat{\bm{\mu}}_j^0\big)^\top\big(\phi_v(\bm{x}_i^0)-\hat{\bm{\mu}}_j^0\big).
\end{split}
\label{eq:MAPest}
\end{equation}

We can update the estimates described in Equation~\ref{eq:MAPest} for $t\ge 1$, which will be explained later in detail. Let $\hat{p}_J^t(\bm{z})$ denotes the estimated GMM at time-step $t$.
Our approach aims to consolidate this distribution in order to maintain model generalization. When we don't have access to source data, we adapt the model such that the encoder consistently aligns the target distributions with this GMM in the embedding space. To achieve this goal, we generate labeled pseudo-data by randomly sampling from the estimated GMM distribution. This pseudo-dataset is denoted as $\mathcal{\hat{D}}^t_{\mathcal{P}}=(\textbf{Z}_{\mathcal{P}}^t,\hat{\textbf{Y}}_{\mathcal{P}}^t)$, where $\bm{Z}_{\mathcal{P}}^t=[\bm{z}_1^t,\ldots,\bm{z}_{N_p}^t]\in\mathbb{R}^{p\times N_p}$, $\hat{\bm{Y}}_{\mathcal{P}}=[\hat{\bm{y}}^{p,t}_1,...,\hat{\bm{y}}^{p,t}_{N_p}]\in \mathbb{R}^{k\times N_p}$, and $\bm{z}_i^t\sim \hat{p}^t_J(\bm{z})$. The labels for these samples are determined based on the model's predictions. However, we only include samples for which the model's prediction confidence level exceeds a predetermined threshold $\tau$, allowing us to exclude outlier samples that lie in the boundary region between modes or in another words, the class clusters.

To ensure that the model retains its generalization power, we augment ERM with domain alignment, and solve the following problem at time step $t$:
\begin{equation}
\footnotesize
\begin{split}
\min_{\bm{v},\bm{w}} \sum_{i=1}^N \mathcal{L}\big(h_{\bm{w}}(\bm{z}_i^p),\hat{\bm{y}}_i^{p,t}\big)+\lambda D\big(\phi_{\bm{v}}(p_t(\bm{X}_t)),\hat{p}_{J}^t(\bm{Z}_{\mathcal{P}}^t)\big).
\end{split}
\label{eq:mainPrMatch}
\end{equation}  
Here, $D(\cdot,\cdot)$ represents a probability discrepancy measure which is our choice. As mentioned before, we use SWD as this metric. The parameter $\lambda$ is a trade-off parameter between the two terms of the optimization.
The first term in the problem formulation aims to ensure that the classifier retains its generalization capability with respect to the internally learned distribution as the model is updated. This term encourages the model to maintain its ability to accurately classify samples based on the internal distribution.
The second term focuses on aligning the target domain distribution with the internally learned distribution within the embedding space. This alignment ensures that the target domain samples are distributed in a manner consistent with the internally learned distribution.
To quantify the discrepancy between distributions, we utilize the SWD metric, enabling us to compare and evaluate the differences between distributions. For a short background on SWD and its definition, please refer to the Appendix.
When we update the model, we can also update the internal distribution $\hat{p}_J^t(\bm{z})$ by incorporating the information from the labeled pseudo-dataset samples. This update process allows us to refine and improve the accuracy of the internally learned distribution based on the newly acquired data.

Solving Eq. \eqref{eq:mainPrMatch} enables the model to achieve good generalization on the new domain $\mathcal{T}^t$. However, catastrophic forgetting remains an unresolved challenge. The reason is that the encoder subnetwork is unconditionally updated to minimize the domain discrepancy term in Eq. \eqref{eq:mainPrMatch}. To address catastrophic forgetting, we utilize  experience replay.
When a task is learned, we select a small subset of the training data points as representative samples, which are then stored in a memory buffer. During model adaptation, these stored samples are replayed to mitigate catastrophic forgetting. Various strategies can be employed for selecting the representative samples, including mean of features (MoF)  \cite{rebuffi2017icarl}, ring buffer  \cite{lopez2017gradient}, and reservoir sampling~ \cite{riemer2018learning}. In our case, since we have access to the internal distribution, MoF becomes a natural choice.

After updating the model at time-step $t$ and subsequently the GMM, we can compute the distance between the representations of all data points in $\mathcal{T}^t$ and their corresponding cluster mean. Specifically, for each data point $\bm{x}^t_l$ such that $\hat{\bm{y}}_l^t=\arg\max f_{\hat{\theta^t}}(\bm{x}^t_l) =j$, we compute the distance as $d^t_{j,l}=|\mu_j^t-\phi(\bm{x}^t_l)|_2^2$. Here, $\mu_j^t$ represents the mean of the $j$-th cluster, and $\phi(\bm{x}^t_l)$ is the representation of $\bm{x}^t_l$ in the embedding space. Given a memory budget of $N_b$ samples, we select $M_b=N_b/k$ samples per class that have the closest distance to the mean. These selected samples form the buffer-stored dataset $\mathcal{D}_b^t=\mathcal{D}_b^{t-1}\cup(\bm{X}_b^t,\hat{\bm{Y}}_b^t)$, where $\bm{X}_b^t$ denotes the selected samples and $\hat{\bm{Y}}_b^t$ represents their corresponding labels.
By including these informative samples, which are the most representative of the data distribution, we update Eq. ~\eqref{eq:mainPrMatch} to tackle catastrophic forgetting. The updated equation is as follows:
\begin{equation}
\small
\begin{split}
&\min_{\bm{v},\bm{w}} \sum_{i=1}^N \mathcal{L}\big(h_{\bm{w}}(\bm{z}_i^p),\hat{\bm{y}}_i^{p,t}\big)+ \sum_{i=1}^{N_b} \mathcal{L}\big(h_{\bm{w}}(\phi_{\bm{v}}(\bm{x}_i^b)),\hat{\bm{y}}_i^{b}\big)\\& + \lambda D\big(\phi_{\bm{v}}(p_t(\bm{X}_t)),\hat{p}_{J}^t(\bm{Z}_{\mathcal{P}}^t)\big)+  \lambda D\big(\phi_{\bm{v}}(p_t(\bm{X}_b^t)),\hat{p}_{J}^t(\bm{Z}_{\mathcal{P}}^t)\big).
\end{split}
\label{eq:mainPrMatchbuffer}
\end{equation}

 \begin{algorithm}[ht!]
\caption{$\mathrm{LDAuCID}\left (\lambda , \tau, N_b\right)$\label{NeuripsUDAalgorithm}} 
 {\small
\begin{algorithmic}[1]
\STATE \textbf{Source Training}: 
\STATE \hspace{2mm}\textbf{Input:} source labeled dataset $\mathcal{D}_{\mathcal{S}}=(\bm{X}_0,  \bm{Y}_0)$
\STATE \hspace{4mm} $\hat{ \theta}_0=(\hat{\bm{w}}_0,\hat{\bm{v}}_0) =\arg\min_{\theta}\sum_i \mathcal{L}(f_{\theta}(\bm{x}_i^0),\bm{y}_i^0)$
\STATE \hspace{2mm}  \textbf{Internal Distribution Estimation:}
\STATE \hspace{4mm} Use Eq. ~\eqref{eq:MAPest} and estimate $\alpha_j^0, \bm{\mu}_j^0,$ and $\Sigma_j^0$
\STATE\hspace{2mm}  \textbf{Memory Buffer Initialization}
\STATE \hspace{4mm} $\mathcal{D}_b^0= (\bm{X}_b^0,\hat{\bm{Y}}_b^0)$ 
 \\ \hspace{4mm} Pick the $N_b/k$ samples with the least \\  \hspace{4mm} $d^t_{j,l}=\|\mu_j^t-\phi(\bm{x}^t_l)\|_2^2$,   $  \hat{\bm{y}}_b^{0,i}=\arg\max f_{\hat{\theta^t}}(\bm{x}_b^{0,N_b}) $
\STATE \textbf{Continual Unsupervised Domain Adaptation}: \FOR{$t=1,\ldots,T$}
\STATE \hspace{2mm} \textbf{Input:} target   unlabeled dataset $\mathcal{D}_{\mathcal{T}}^t=(\bm{X}_t)$
\STATE \hspace{2mm} \textbf{Pseudo-Dataset Generation:} 
\STATE \hspace{2mm} $\mathcal{\hat{D}}_{\mathcal{P}}^t=(\textbf{Z}_{\mathcal{P}}^t,\hat{\textbf{Y}}_{\mathcal{P}}^t)=$
\STATE \hspace{3mm} $([\bm{z}_1^{p,t},\ldots,\bm{z}_N^{p,t}],[\hat{\bm{y}}_1^{p,t},\ldots,\hat{\bm{y}}_N^{p,t}])$, where:
   $\bm{z}_i^{p,t}\sim \hat{p}^{t-1}_J(\bm{z}), 1\le i\le N_p$ \hspace{2mm} and
\\ \hspace{4mm}$\hat{\bm{y}}_i^{p,t}=
\arg\max_j\{h_{\hat{\bm{w}}_t}(\bm{z}_i^{p,t})\}$ if with confidence $\tau$: $\max_j\{h_{\hat{\bm{w}}_t}(\bm{z}_i^{p,t})\}>\tau$
\FOR{$itr = 1,\ldots, ITR$ }
\STATE draw data batches from $\mathcal{D}_{\mathcal{T}}^t$ and $\mathcal{\hat{D}}_{\mathcal{P}}$
\STATE Update the model by solving Eq. ~\eqref{eq:mainPrMatchbuffer}
\ENDFOR
\STATE    \textbf{Internal Distribution Estimate Update:} 
\STATE \hspace{6mm} Use Eq. ~\eqref{eq:MAPest} similar to step 5 above.
\STATE \textbf{Memory Buffer Update}
\STATE \hspace{6mm} $\mathcal{D}_b^t=\mathcal{D}_b^{t-1}\cup(\bm{X}_b^t,\hat{\bm{Y}}_b^t)$, where $(\bm{X}_b^t,\hat{\bm{Y}}_b^t)$   is   computed  similar to step 7 above. 
\ENDFOR
\end{algorithmic}}
\end{algorithm}

The inclusion of the second supervised term in Eq. \eqref{eq:mainPrMatchbuffer} serves to alleviate catastrophic forgetting. By incorporating this term, we ensure that the model retains knowledge from previous tasks while adapting to learn new ones. Additionally, the fourth term in Eq. \eqref{eq:mainPrMatchbuffer} plays a crucial role in consolidating the internal distribution across all the previously encountered tasks. This consolidation step enhances the model's ability to generalize across various domains.
Our algorithm called Continual Domain Adaptation Using Consolidated Internal Distribution (LDAuCID) is presented and outlined in Algorithm~\ref{NeuripsUDAalgorithm}. This algorithm provides a step-by-step procedure for adapting the model to new tasks while mitigating catastrophic forgetting and consolidating the internal distribution.
Furthermore, Figure~\ref{figDL:LL} visually illustrates the LDAuCID framework, showcasing the key components and processes involved in continual domain adaptation.

   \section{Theoretical Analysis}

We present an analysis of our approach within  a probably approximately correct learning (PAC-Learning) formulation  \cite{shalev2014understanding}. In short, we prove that at each time step, our algorithm minimizes and upperbound for the expected error on all tasks learned so far. As a result, our method enables knowledge transfer and at the same time mitigates forgetting effects. 

To formalize our analysis, we consider the shared embedding space to be our input space. We then set the hypothesis space to be the set of classifier sub-networks denoted as $\mathcal{H} = \{h_{\bm{w}}(\cdot) \mid h_{\bm{w}}(\cdot): \mathcal{Z} \rightarrow \mathbb{R}^k, \bm{v} \in \mathbb{R}^V\}$, where each $h_{\bm{w}}(\cdot)$ maps an input from the input space $\mathcal{Z}$ to a $k$-dimensional output space, and $\bm{w}$ represents the parameters of the sub-network.
In this context, let $e_0$ represent the expected error on the source domain, which is the true error that the classifier sub-network incurs when classifying samples from the source domain. Similarly, $e_t$ denotes the expected error on the target domains. Additionally, we define $e_{t}^J$ as the expected error on the pseudo-dataset for a given classifier sub-network $h \in \mathcal{H}$. The pseudo-dataset represents the labeled samples generated by the consolidated distribution.

To characterize the empirical distributions in the embedding space, we define $\hat{p}_0(\bm{x})$ as the empirical source distribution, which is estimated by computing the average representation of the source domain samples $\bm{x}_n^s$ using the embedding function $\phi_{\bm{v}}(\cdot)$. Similarly, $\hat{p}_t(\bm{x})$ represents the empirical target distribution in the embedding space for subsequent tasks.
Based on these definitions, we provide the following theorem for our algorithm.


\textbf{Theorem 1}: Consider LDAuCID algorithm at learning time-step $t=T$. Then for all the previously learned tasks $t<T$, the following holds:
\begin{equation}
\small
\begin{split}
&e_{t}\le  e_{T-1}^J +\alpha SW(\phi(\hat{p}^{t}), \hat{p}_{J}^{t})^\beta+\alpha\sum_{s=t}^{T-2} W(\hat{p}_{J}^{s},\hat{p}_{J}^{s+1})^\beta\\& +e(\bm{w}^*) +\sqrt{\big(2\log(\frac{1}{\xi})/\zeta\big)}\big(\sqrt{\frac{1}{M_t}}+\sqrt{\frac{1}{N_p}} +\sqrt{\frac{1}{N_b}}\big),
\end{split}
\label{eq:theroemfromcourtyCatForoursDAL}
\end{equation} 
where $e(\bm{w}^*)$ denotes expected error for the   joint-trained optimal model in the hypothesis space, i.e., the model trained on both the internal distribution and the current task data: $\bm{w}^*= \arg\min_{\bm{w}} e_{c}(\bm{w})=\arg\min_{\bm{w}}\{ e_{t}(h)+  e_{J}^t(h)\}$,   $SW(\cdot,\cdot)$ denotes the WD distance, and $\zeta$, $\xi$, and $\alpha$ are    constants and $\beta=(2(d+1))^{-1}$.  Finally, $M_t$, $N_p$, and $N_b$ denote the size of the current task training dataset, the pseudo-dataset, the memory buffer samples.

\textbf{Proof:}  We base our analysis   on the following theorem~ \cite{redko2017theoretical} which is derived for the classic  UDA when there is a single source domain and a target domain, and data points for both domains are accessible simultaneously:

 \textbf{Theorem~2:}
 Suppose we have two classification tasks occurring in two distinct domains, each characterized by probability distributions denoted as $p^t$ and $p^{t'}$. We consider training datasets containing $n_t$ and $n_{t'}$ data points for these tasks, respectively. Let $h_{w^{t'}}$ represent a model trained specifically for the source domain. Now, assuming that $d'>d$ and $\zeta<\sqrt{2}$, we can establish the existence of a constant $N_0$  which is dependent on the value of $d'$ such that for any $\xi>0$  when  $\min(n_t, n_{t'})\ge N_0\max(\xi^{-(d'+2)},1)$, the following statement holds true for all models $h$ with a probability of at least $1-\xi`$:
\begin{equation}
\begin{split}
&e_{t}(h)  \le e_{t'}(h) +  W(\hat{p}^{t}, \hat{p}^{t'})+ e(\bm{w}^*)+ \sqrt{\big(2\log(\frac{1}{\xi})/\zeta\big)}\big(\sqrt{\frac{1}{n_t}}+\sqrt{\frac{1}{n_{t'}}}\big).
\end{split}
\label{eq:theroemfromcourtyCatForDAL}
\end{equation}  
Here, $e(\bm{w}^*)$ is the expected error for the optimal model obtained by joint training.
 
 Theorem~2   provides an upper-bound estimation on the performance of a particular model that has been trained in a source domain when it is used in a target domain. This theorem is symmetric, meaning that the two domains can be interchanged or shuffled. The theorem suggests that if the difference between the probability distributions of the two domains, as measured by the Wasserstein distance, is small, and if the joint-trained optimal model performs well with low expected error, then the performance of the model in the target domain will be similar to its performance in the source domain.
Additionally, the third term of the theorem indicates that in order for domain adaptation to be possible, the base model should have the ability to learn both tasks jointly. This means that the model needs to be capable of acquiring knowledge from both the source and target domains simultaneously. For example, consider the ``XOR classification problem'' in binary classification. It has been shown that a single model cannot effectively learn both tasks in this case ~ \cite{mangal2007analysis}. Therefore, it is necessary for the two domains to be relevant and compatible with each other in terms of the tasks they involve.

 We also rely  on the following inequality~ \cite{bonnotte2013unidimensional}  to extend Theorem~2 to the case of using SWD  for domain alignment:\begin{eqnarray}
SW(p_X,p_Y)\leq W(p_X,p_Y) \leq \alpha SW(p_X,p_Y)^\beta.
\label{eq:inequalities}
\end{eqnarray}
 We use the above two results~ \cite{bonnotte2013unidimensional,redko2017theoretical} to prove Theorem 1.

In Theorem 2, we consider a specific case where the distribution  $\phi(p^{t})$ serves as the target domain. Additionally, we consider the pseudo-task with the distribution $p^{J}{T-1}$ as the source domain within the embedding space. To establish the result, we use the  triangular inequality recursively on the term $W(\phi(\hat{p}^{t}), \hat{p}^{J}{T-1})$, as stated in Eq.  \eqref{eq:theroemfromcourtyCatForDAL}:
\begin{equation}
W(\phi(\hat{p}^{t}), \hat{p}^{J}{s}) \le W(\phi(\hat{p}^{t}), \hat{p}^{J}{s-1}) + W(\hat{p}^{J}{s}, \hat{p}^{J}{s-1}),
\end{equation}
This inequality is utilized iteratively for all time steps, where $t \leq s < T$. By summing up all the resulting terms, and applying Eq. ~\eqref{eq:inequalities} to replace the WD terms with SWD terms, we arrive at the desired conclusion presented in Eq.  \eqref{eq:theroemfromcourtyCatForoursDAL}.

The effectiveness of the LDAuCID algorithm can be explained by Theorem 1. The major terms on the right-hand side of Eq.  \ref{eq:theroemfromcourtyCatForoursDAL}, which serve as an upper bound for the expected error of each  domain, are continually minimized by LDAuCID.
The first term is minimized because the algorithm utilizes random samples from the internal distribution to minimize the empirical error term for the estimation of the internal distribution, as represented by the first term of Eq.  \eqref{eq:mainPrMatchbuffer}.
The second term is minimized through the alignment of the task distribution with the empirical internal distribution in the embedding space at time $t$, as indicated by the third term of Equation \eqref{eq:mainPrMatchbuffer}.
The third term, which is a summation capturing the effect of continual learning, is minimized when each task $\mathcal{T}^s$ is learned and the internal distribution is deliberately updated. We can also conclude that a GMM should be a good estimate for the internal distribution for our method to work. We don't impose a constraint to impose this result but if the model is complex enough to learn the domains, we expect that a multimodal internal distribution is learned. This additive term grows as more tasks are learned after a particular task, potentially making the upperbound looser and leading to increased forgetting effects.
More forgetting on tasks that are learned earlier is intuitive.
The fourth term is a small constant when the tasks are related and share the same classes with the same label space. LDAuCID does not directly minimize this term. However, it suggests that the model should perform well in a joint-training scenario in order to exhibit good performance in a sequential learning setting.
The last term is a constant term that becomes negligible when an adequate number of source and target training data points are available. It also implies that storing more samples in the memory buffer leads to improved performance, as expected.
In conclusion, if the upper bound in Eq.  \eqref{eq:theroemfromcourtyCatForoursDAL} is sufficiently tight, the domains are relevant, and the hypothesis space is suitable for learning the tasks, then adapting the model using LDAuCID can effectively address catastrophic forgetting and enhance model generalization on the target domains. Hence LDAuCID is able to address the major challenges of UDA and CL simultaneously.

 \section{Empirical Validation}
We first explain the setup used in our experiments and then  provide our results.    Our implemented code is accessible at \url{https://github.com/rostami-m/LDACID}. 

   \subsection{Experimental Setup}

 \subsubsection{Datasets and Tasks}

We leverage four well-established UDA benchmark datasets and adapt them for UDA within a CL setting. We modify these datasets to create sequential UDA tasks, allowing us to validate our approach on these widely recognized datasets.
To ensure fair comparison with existing UDA methods, we conduct experiments using the same learning setting employed in those methods, where only one source domain and one target domain are considered. We adhere to the evaluation protocols commonly used in recent UDA papers to ensure consistency and enable meaningful comparisons with other approaches in the field.

\textbf{Digit recognition   tasks:}    three commonly used datasets are utilize in this benchmark, namely MNIST ($\mathcal{M}$), USPS ($\mathcal{U}$), and SVHN ($\mathcal{S}$), to represent three distinct domains. Most existing UDA methods report their results on three specific tasks involving these domains: $\mathcal{M}\rightarrow \mathcal{U}$, $\mathcal{U}\rightarrow \mathcal{M}$, and $\mathcal{S}\rightarrow \mathcal{M}$. We conduct experiments on six possible task orders on these three tasks. By including these sequential tasks, we cover all three of the classic UDA tasks and provide a more extensive assessment of our method's performance.
We adjusted the dimensions of the images in the SVHN dataset to be $28\times 28$ pixels in size. This resizing was necessary to ensure consistency in the input data across all domains, as we employed the same encoder for all domains.  

\textbf{ImageCLEF-DA Dataset:} this benchmark   consists of 12 common image classes shared by three visual recognition datasets: Caltech-256 ($\mathcal{C}$), ILSVRC 2012 ($\mathcal{I}$), and Pascal VOC 2012 ($\mathcal{P}$). The benchmark is   balanced, with each class having an equal representation of 50 images, resulting in a total of 600 images for each domain.
There are six possible binary UDA tasks that can be defined. To comprehensively evaluate our method, we conduct experiments on all possible ternary sequential UDA tasks. By including these ternary tasks, we cover all six classic binary UDA tasks, allowing for a thorough assessment of our method's performance across various domain transitions.

\textbf{Office-Home Dataset}: This benchmark dataset presents a more challenging object recognition scenario, consisting of a collection of 15,500 images captured in office and home settings. The images are categorized into 65 distinct classes, encompassing a wide range of objects and scenes. The dataset comprises four domains, each exhibiting significant variations and gaps: Artistic images ($A$), Clip Art ($C$), Product images ($P$), and Real-World images ($R$). These domain disparities provide the opportunity to define 12 pair-wise binary Unsupervised Domain Adaptation (UDA) tasks. The challenging nature of this benchmark dataset, along with the distinct gaps between the domains, provides an ideal testbed to evaluate the effectiveness of our approach in addressing UDA challenges. 
To thoroughly evaluate our method's capabilities, we conduct experiments on 24 possible sequential UDA tasks. By considering these sequential tasks, we cover the full spectrum of domain transitions, allowing for a comprehensive assessment of our method's performance.

\textbf{Office-Caltech Dataset}: This benchmark for object recognition focuses on a specific set of 10 classes that are shared between the Office-31 and Caltech-256 datasets. The dataset encompasses four distinct visual domains: Amazon ($A$), Caltech ($C$), DSLR ($D$), and Webcam ($W$). In total, the dataset consists of 2533 images, distributed across these four domains.
Within this benchmark, there are 12 binary UDA tasks that can be defined. To comprehensively evaluate our method, we conduct experiments on 24 sequential tasks. By considering these sequential tasks, we cover a wide range of domain transitions, allowing us to thoroughly assess the performance of our method.

For our experiments, we employed cross entropy loss as the discrimination loss. During each training epoch, we calculated the combined loss function using the training data and terminated the training process once the loss function reached a relatively constant value. We utilized the Keras   for implementation and employed the ADAM optimizer. 
The code implementation was executed on a cluster node equipped with two Nvidia Tesla P100-SXM2 GPUs, allowing for efficient parallel processing. 
Each dataset in all domains has its predefined standard training and testing splits. In our experiments, we employed these testing splits to evaluate the performance of the methods  in terms of classification accuracy. 
To ensure robustness and reliability of our results, we conducted five training trials for each experiment. We then reported the average performance and the standard deviation of these trials on the respective testing sets.  

\subsubsection{Network Structure and Evaluation Protocol}

 In our  experiments, we followed the precedent to select the specific neural network architectures to serve as the base models. For the digit recognition tasks, we utilized the VGG16 network as the base model. We employed the Decaf6 features for the Office-Caltech tasks. On the other hand, for the remaining two datasets, we opted for the ResNet-50 network, which had been pre-trained on the   ImageNet dataset, to serve as the backbone encoder.
To gain insights into the learning process of the models over time, we generated learning curves. These curves illustrate the performance of the model on the testing split of the learned tasks as a function of the training epochs. By examining the learning curves, we were able to study the dynamics of learning, observing how the model's performance evolved throughout the training process. Our aim was to simulate continual training during the execution phase, allowing us to monitor and analyze the learning dynamics.
During the training process, we trained the base model using the labeled data from the source domain. We then evaluated the performance of the model before adaptation, serving as a simple ablation study, to assess the impact of domain shift. Subsequently, we applied the LDAuCID algorithm to adapt the model using the unlabeled data from the target domain. We reported the performance of LDAuCID on the target domains.

\subsubsection{Baselines for Comparison}

In order to evaluate the effectiveness of our LDAuCID algorithm, we conducted comparative analyses against existing UDA methods. We considered our algorithm in the scenario where only two domains are present, allowing us to compare its performance against classic UDA algorithms. In this case, we focused on a single target domain and reported our algorithm's performance specifically on that domain. While our approach still addresses UDA in a sequential learning setting, we acknowledge that this is the most similar   framework available for the purpose of comparison for our study.

For the purpose of comparison, we included several well-known classic UDA methods in our evaluation. These methods include GtA  \cite{sankaranarayanan2018generate}, DANN  \cite{ganin2016domain}, ADDA  \cite{tzeng2017adversarial}, MADA  \cite{pei2018multi}, DAN  \cite{long2015learning}, DRCN  \cite{ghifary2016deep}, RevGrad  \cite{ganin2014unsupervised}, JAN  \cite{long2017deep}, JDDA  \cite{chen2019joint}, and UDAwSD  \cite{li2020model}. We made use of the reported results from these works, provided they were available for the corresponding datasets. These methods include both methods based on adversarial learning as well as direct probability matching. 

In the tables presented in our study, we used bold font to indicate the best performance achieved in each case. This allowed for easy identification and comparison of the different methods and their respective results.

\subsection{Results}

 We first provide the learning curves for our algorithm to study the temporal aspects of learning and then focus on comparison against existing UDA methods.
\subsubsection{Dynamics of Learning}

\begin{figure*}[tb!]
    \centering
           \begin{subfigure}[b]{0.32\textwidth}\includegraphics[width=\textwidth]{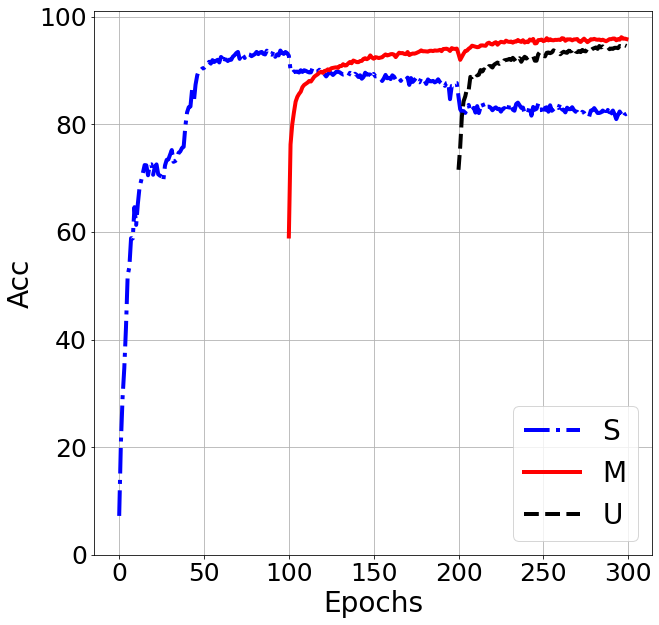}
           \centering
        \caption{$\mathcal{S}\rightarrow\mathcal{M}\rightarrow \mathcal{U}$}
        \label{NIPSDALfig:Digits1}
    \end{subfigure}
  \centering
           \begin{subfigure}[b]{0.32\textwidth}\includegraphics[width=\textwidth]{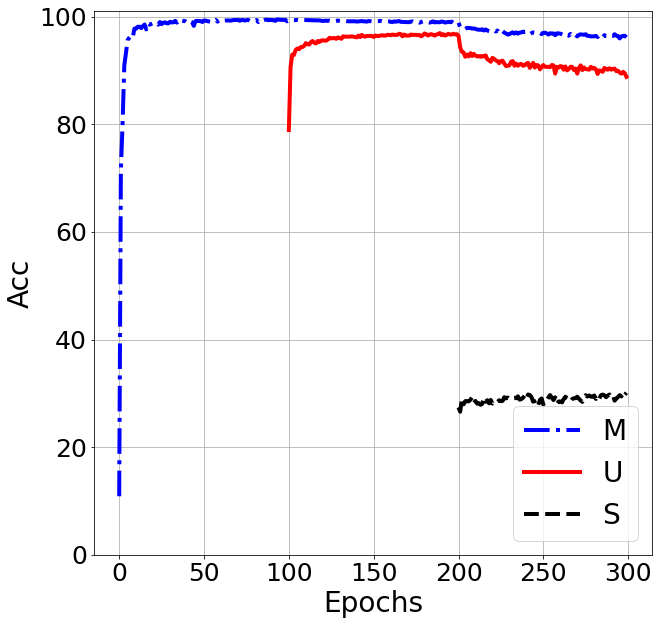}
           \centering
        \caption{$\mathcal{M}\rightarrow\mathcal{U}\rightarrow \mathcal{S}$}
        \label{NIPSDALfig:Digits2}
    \end{subfigure}
      \centering
           \begin{subfigure}[b]{0.32\textwidth}\includegraphics[width=\textwidth]{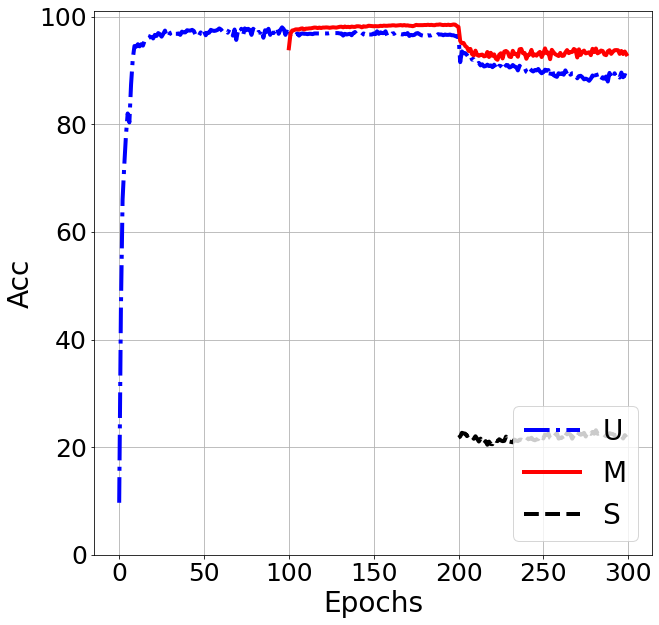}
           \centering
        \caption{$\mathcal{U}\rightarrow\mathcal{M}\rightarrow \mathcal{S}$}
        \label{NIPSDALfig:Digits3}
    \end{subfigure}
\\
    \begin{subfigure}[b]{0.32\textwidth}\includegraphics[width=\textwidth]{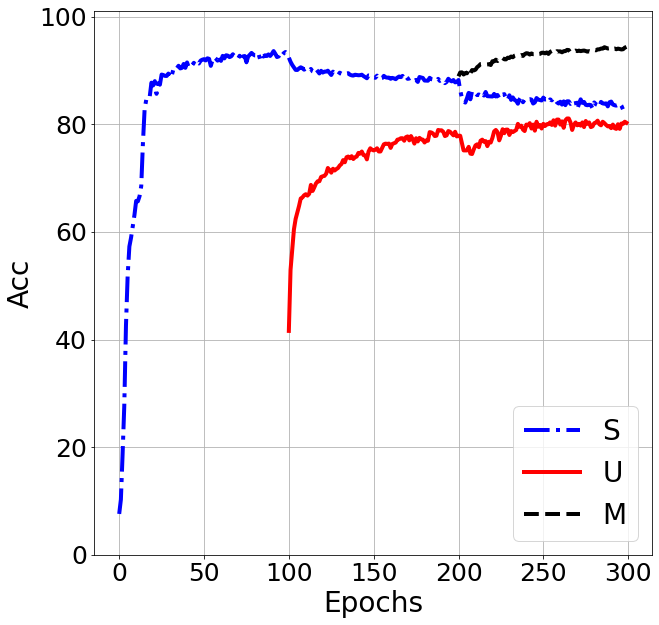}
           \centering
        \caption{$\mathcal{S}\rightarrow\mathcal{U}\rightarrow \mathcal{M}$}
        \label{NIPSDALfig:Digits4}
    \end{subfigure}
      \centering
           \begin{subfigure}[b]{0.32\textwidth}\includegraphics[width=\textwidth]{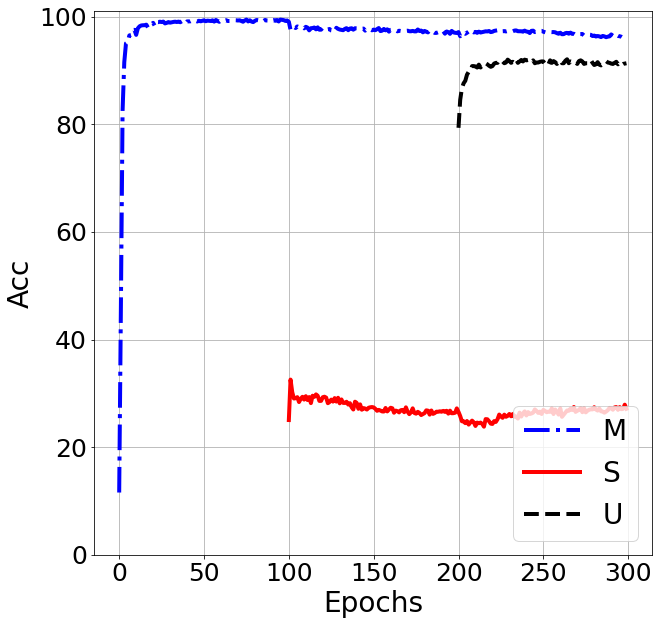}
           \centering
        \caption{$\mathcal{M}\rightarrow\mathcal{S}\rightarrow \mathcal{U}$}
        \label{NIPSDALfig:Digits5}
    \end{subfigure}
      \centering
           \begin{subfigure}[b]{0.32\textwidth}\includegraphics[width=\textwidth]{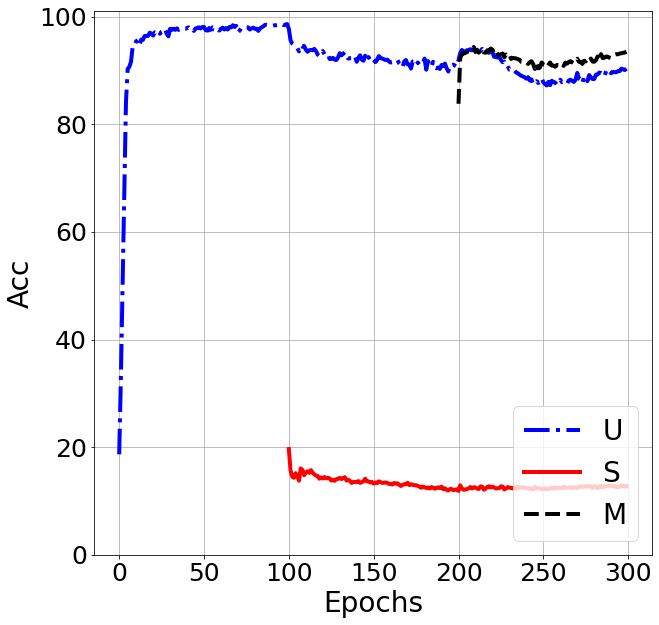}
           \centering
        \caption{$\mathcal{U}\rightarrow\mathcal{S}\rightarrow \mathcal{M}$}
        \label{NIPSDALfig:Digits6}
    \end{subfigure}
     \caption{Learning curves for sequential UDA tasks on  digit benchmark. (Best viewed in color).  }\label{NIPSDALfig:contrelating1}
\end{figure*}

Figure~\ref{NIPSDALfig:contrelating1} depicts the learning curves for the digit recognition sequential UDA tasks. Each curve represents the model's performance on a specific task as it progresses through 100 training epochs. After completing one task, the model moves on to learn the subsequent task. Throughout the training process, we have maintained a memory buffer containing 10 samples per class for each domain, which facilitates experience replay and addressing catastrophic forgetting.

A major observation that we have is that when we learn $\mathcal{S}$ after either $\mathcal{M}$ or $\mathcal{U}$ datasets, the model performance on the SVHN dataset is very low, around 20\%. This is an expected observation and it is exactly the reason behind not using the SVHN dataset as the source model in the UDA literature for MNIST or USPS datasets. SVHN is a more complex dataset and when a model is fine-tuned on $\mathcal{M}$ or $\mathcal{U}$ datasets, its generalizability on more complex datasets is reduced significantly. We conclude that in order to succeed in UDA, the source domain should be at least as complex as the target domains, which is the trend in the literature. Note, however, in all tasks, we address forget effects quite well.

An interesting observation from the learning curves is that the initial drop in performance on the previously learned  domains upon encountering domain shift in both tasks. For example, we observe at Figure~\ref{NIPSDALfig:Digits1} that after the 200$^{th}$ epoch, performances for both $\mathcal{S}$ and $\mathcal{M}$ datasets drop which indicates minor forgetting effects. But after the initial drop, the performance remains relatively stable and in the case of $\mathcal{M}$, we even see performance improvement. The initial performance drop on older tasks is because when we start learning a new task, the model's learnable parameters deviate from previous values. However, when we benefit from experience replay, the performance is maintained. If the current task is similar to a previous task, e.g., $\mathcal{S}$ and $\mathcal{M}$, transfer learning can help performance improvement on the past task. This improvement is consistently observed across   target domain tasks, indicating that LDAuCID successfully leverages past experiences to enhance the model's capabilities. We also observe a notable jumpstart in performance for all subsequent target tasks after the initial task. In other words, the learning curves for the subsequent target tasks start at significantly higher testing accuracies than what would be expected from random label assignment. This initial jumpstart can be attributed to the presence of shared similarities between the domains, which allows for knowledge transfer from previous experiences. For instance, in Figure~\ref{NIPSDALfig:Digits1}, we observe that the initial testing accuracies for domains $\mathcal{M}$ and $\mathcal{U}$ are both in the range of approximately 60\%.

Another important observation is the effective mitigation of catastrophic forgetting. As we progress to learn the subsequent domains for all six tasks, we notice that the model's performance on the previously learned tasks remains relatively stable, indicating a minimal forgetting effect. This onservation implies that the knowledge gained from earlier tasks is retained to a great extent and is not   overwritten significantly by the new learning experiences.
It is worth mentioning that the SVHN dataset in the $\mathcal{S}\rightarrow\mathcal{U}\rightarrow \mathcal{M}$ and $\mathcal{S}\rightarrow\mathcal{M}\rightarrow \mathcal{U}$ tasks exhibits slightly more severe forgetting effects compared to the other datasets for the same reason we initially provided. Due to the higher difficulty level of the SVHN dataset, we may require a larger number of samples in the memory buffer for  mitigation of forgetting to a degree similar to simpler datasets. Nonetheless, even in these more challenging scenarios, the LDAuCID algorithm demonstrates its ability to improve model generalization while effectively mitigating catastrophic forgetting.
Overall, based on these observations, we can conclude that the LDAuCID algorithm is successful in enhancing the model's performance and generalization capabilities while effectively managing the detrimental effects of catastrophic forgetting in these sequential UDA tasks.

 \begin{figure*}[tb!]
    \centering
          \begin{subfigure}[b]{0.32\textwidth}\includegraphics[width=\textwidth]{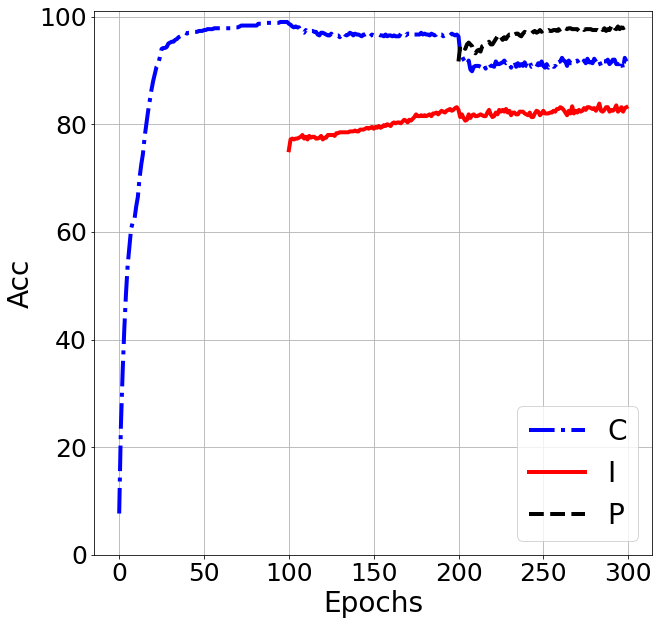}
           \centering
        \caption{$\mathcal{C}\rightarrow \mathcal{I}\rightarrow \mathcal{P}$ }
        \label{NIPSDALfig:imageclef1}
    \end{subfigure}
              \begin{subfigure}[b]{0.32\textwidth}\includegraphics[width=\textwidth]{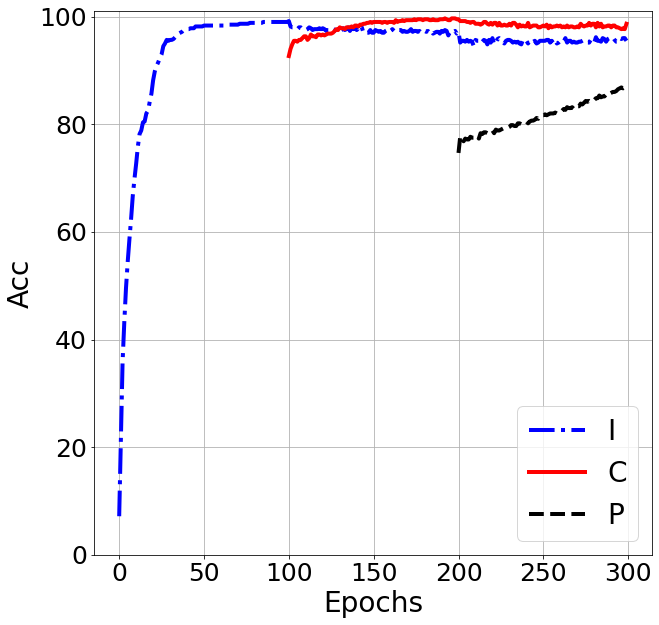}
           \centering
        \caption{$\mathcal{I}\rightarrow \mathcal{C}\rightarrow \mathcal{P}$ }
        \label{NIPSDALfig:imageclef2}
    \end{subfigure}
              \begin{subfigure}[b]{0.32\textwidth}\includegraphics[width=\textwidth]{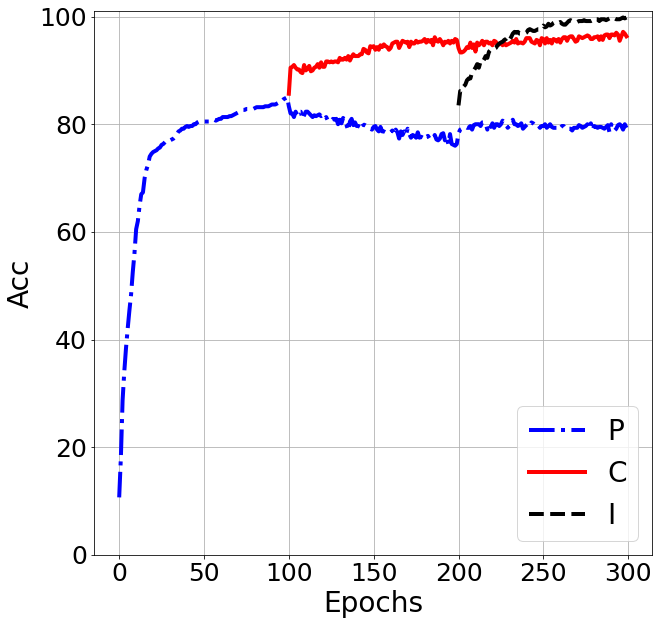}
           \centering
        \caption{$\mathcal{P}\rightarrow \mathcal{C}\rightarrow \mathcal{I}$ }
        \label{NIPSDALfig:imageclef3}
    \end{subfigure}
    \\
       \begin{subfigure}[b]{0.32\textwidth}\includegraphics[width=\textwidth]{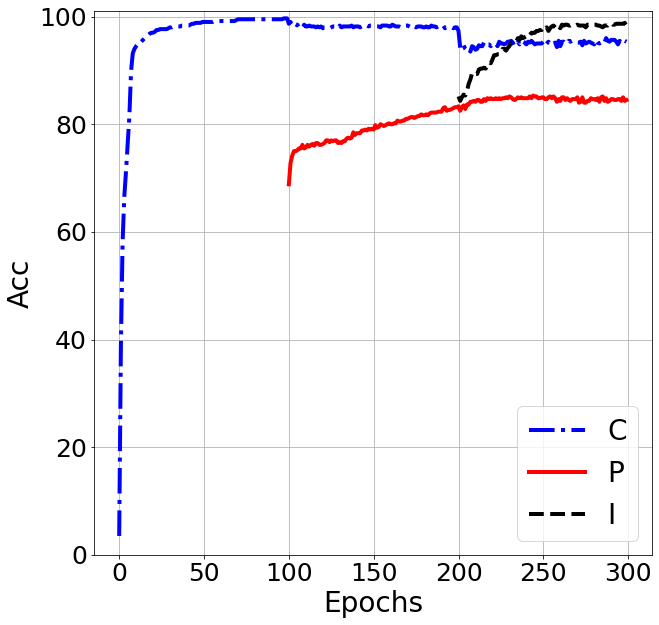}
              \centering
        \caption{$\mathcal{C}\rightarrow \mathcal{P}\rightarrow \mathcal{I}$}
        \label{NIPSDALfig:imageclef4}
    \end{subfigure}
              \begin{subfigure}[b]{0.32\textwidth}\includegraphics[width=\textwidth]{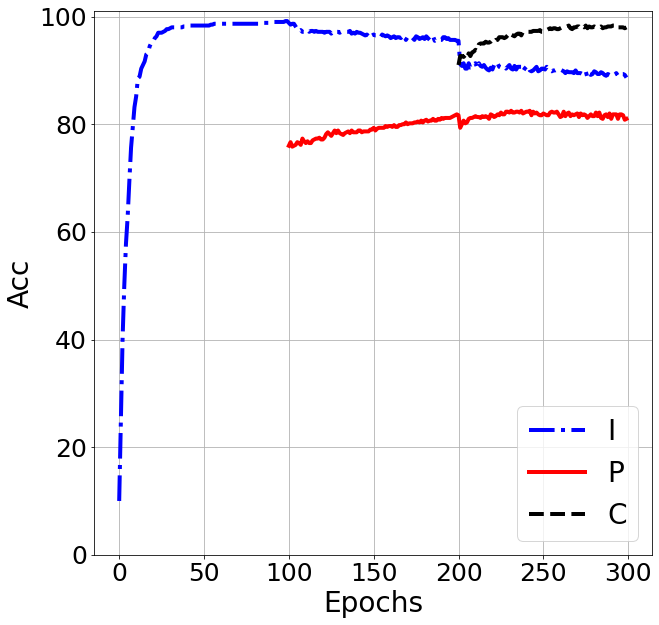}
           \centering
        \caption{$\mathcal{I}\rightarrow \mathcal{P}\rightarrow \mathcal{C}$ }
        \label{NIPSDALfig:imageclef5}
    \end{subfigure}
              \begin{subfigure}[b]{0.32\textwidth}\includegraphics[width=\textwidth]{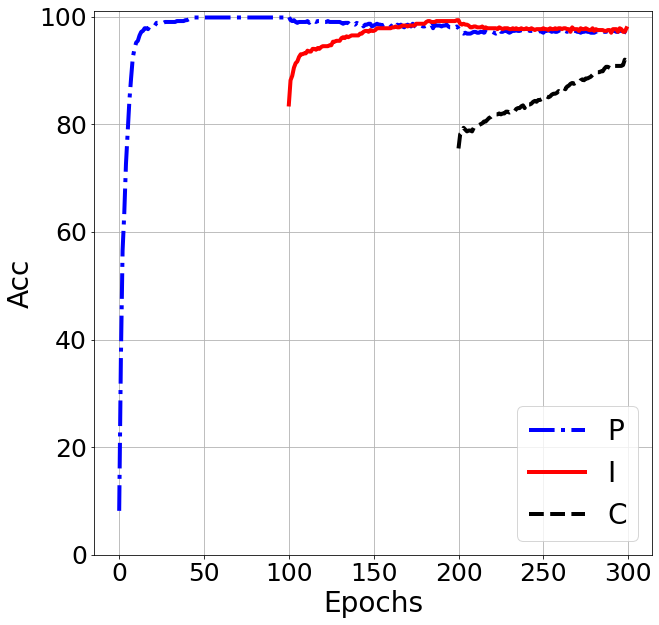}
           \centering
        \caption{$\mathcal{P}\rightarrow \mathcal{I}\rightarrow \mathcal{C}$ }
        \label{NIPSDALfig:imageclef6}
    \end{subfigure}
     \caption{Learning curves for sequential UDA tasks on   ImageClef. (Best viewed in color).  }\label{NIPSDALfig:contrelating2}
\end{figure*}

The learning curves for the six sequential UDA tasks of the ImageCLef dataset are presented in Figure \ref{NIPSDALfig:contrelating2}, showcasing a similar behavior to the previously discussed Figure \ref{NIPSDALfig:contrelating1}. Once again, we observe an initial drop in performance on the past learned domains due to domain shift in all sequential tasks, which is followed by a stable performance on the previously learned domains when a new domain is learned. This observation indicates that we address catastrophic forgetting in all tasks quite well.
We also see a significant jumpstart performance for all tasks because in all cases, the initial perofrmnace is around \%80.
This observation indicates that there is knowledge transfer from previous experiences, resulting in a higher initial testing accuracy compared to random label assignment. The rising trend in the learning curves after the initial phase signifies the effectiveness of the LDAuCID algorithm in enhancing the performance of the source-trained model on the target domains due to distribution alignment and at the same time maintaining generalizability on past learned tasks using experience replay.

Interestingly, we observe notable improvements in performance for previously learned domains in certain cases. For example, we can observe in Figure \ref{NIPSDALfig:imageclef1} that the performance of domain $\mathcal{I}$ increases after learning domain $\mathcal{P}$ as the third task. This finding suggests that knowledge transfer can occur not only from past domains to the current one, but also from the current domain to previously learned domains. This trend implies that leveraging past experiences can not only enhance the learning of the current task but also improve performance on tasks encountered earlier.
When comparing Figure \ref{NIPSDALfig:imageclef1} and Figure \ref{NIPSDALfig:imageclef6}, we can also observe a notable distinction in the asymptotic performance of the second task, $\mathcal{I}$, in the sequential learning curves. This observation leads to an intriguing conclusion that the order in which tasks are learned can have a significant impact on the overall performance in a continual learning   setting. While we may not always have direct control over the task order in CL scenarios, this finding highlights a potential research direction. It prompts us to explore the optimization of task order in settings where such choices can be made.
By investigating the optimal task order for improved performance, we could potentially enhance the learning outcomes in sequential UDA. This research direction becomes particularly relevant when the agent has the ability to select which task to learn next, extending the concept of active learning to continual learning. By strategically determining the task order, we can potentially maximize the benefits of knowledge transfer and improve overall performance in continual learning settings. Moreover, we conclude that an optimal CL algorithm should perform well in all possible task orders.

 \begin{figure*} 
    \centering
           \begin{subfigure}[b]{0.22\textwidth}\includegraphics[width=\textwidth]{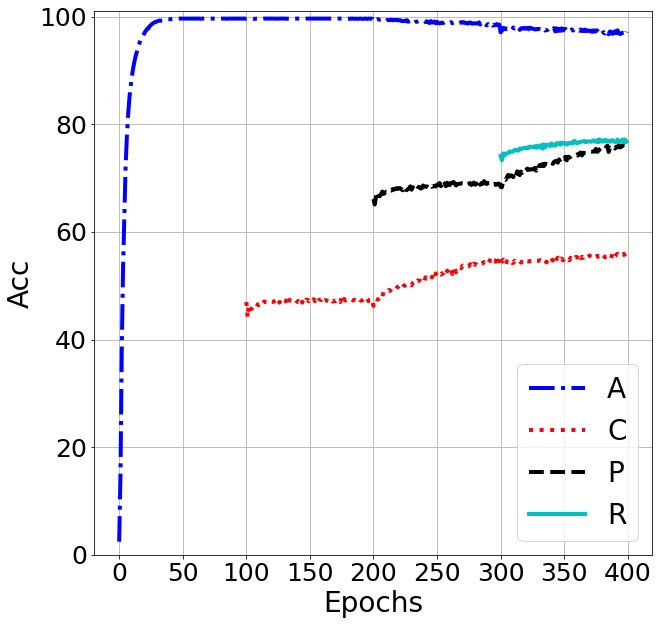}
           \centering
        \caption{$A \rightarrow C \rightarrow P \rightarrow  R  $}
        \label{NIPSDALfig:officehome1}
    \end{subfigure}
    \begin{subfigure}[b]{0.22\textwidth}\includegraphics[width=\textwidth]{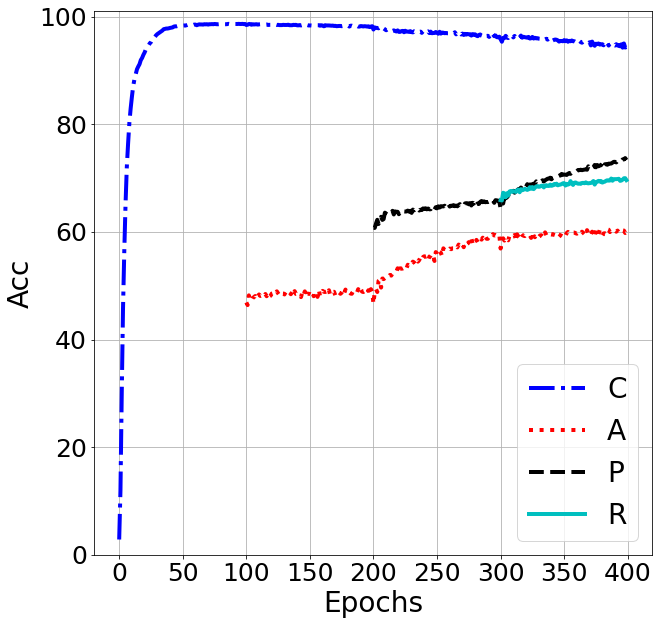}
           \centering
        \caption{$C \rightarrow A \rightarrow P \rightarrow  R  $}
        \label{NIPSDALfig:officehome2}
    \end{subfigure}
               \begin{subfigure}[b]{0.22\textwidth}\includegraphics[width=\textwidth]{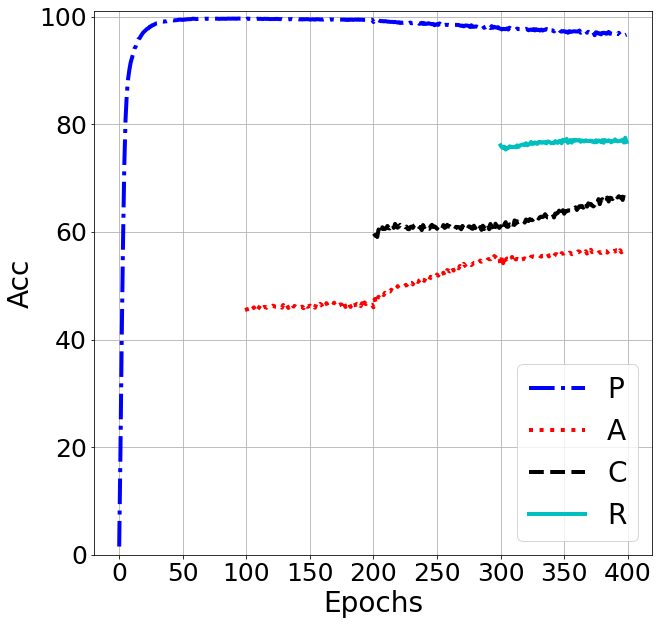}
           \centering
        \caption{$P \rightarrow A \rightarrow C \rightarrow  R  $}
        \label{NIPSDALfig:officehome3}
    \end{subfigure}
    \begin{subfigure}[b]{0.22\textwidth}\includegraphics[width=\textwidth]{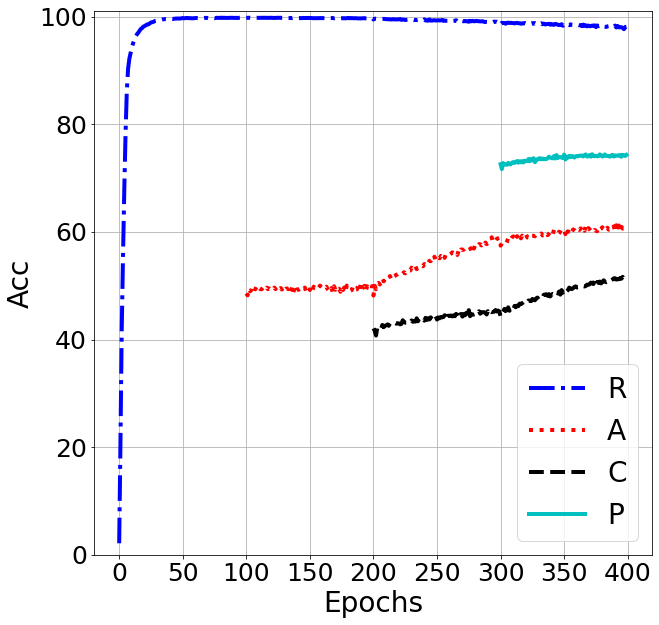}
           \centering
        \caption{$R \rightarrow A \rightarrow C \rightarrow  P  $}
        \label{NIPSDALfig:officehome4}
    \end{subfigure}\\    \centering
           \begin{subfigure}[b]{0.22\textwidth}\includegraphics[width=\textwidth]{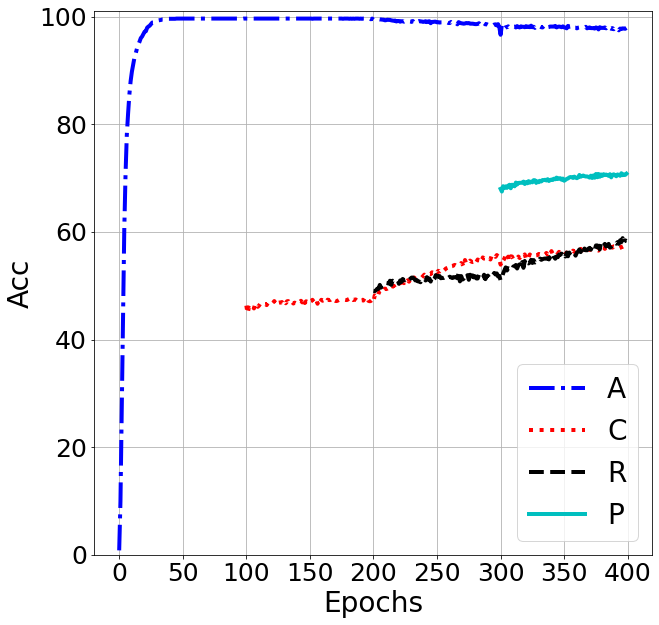}
           \centering
        \caption{$A \rightarrow C \rightarrow R \rightarrow  P  $}
        \label{NIPSDALfig:officehome5}
    \end{subfigure}
    \begin{subfigure}[b]{0.22\textwidth}\includegraphics[width=\textwidth]{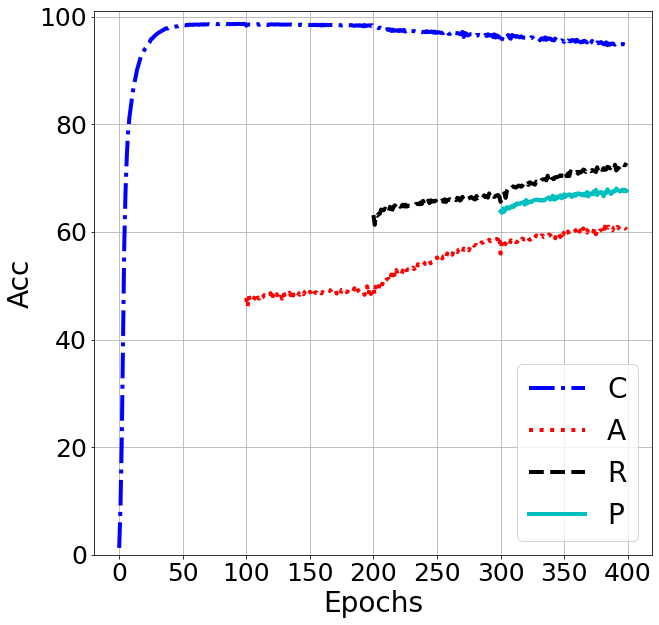}
           \centering
        \caption{$C \rightarrow A \rightarrow R \rightarrow  P  $}
        \label{NIPSDALfig:officehome6}
    \end{subfigure}
               \begin{subfigure}[b]{0.22\textwidth}\includegraphics[width=\textwidth]{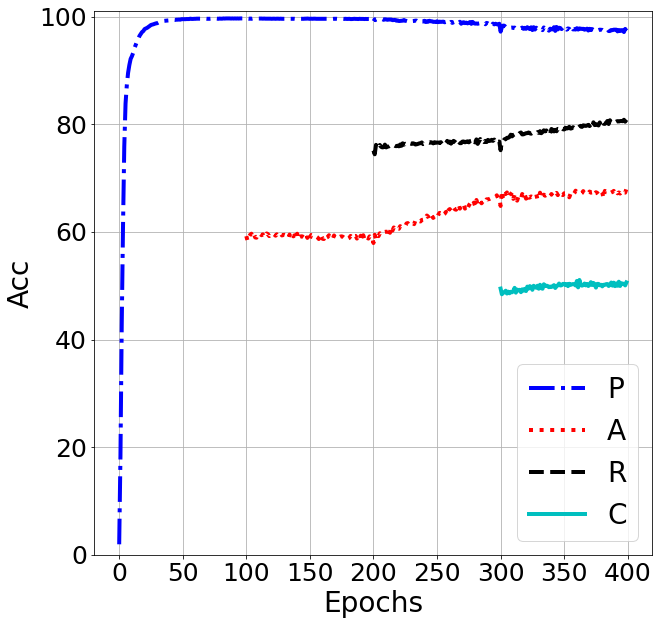}
           \centering
        \caption{$P \rightarrow A \rightarrow R \rightarrow  C  $}
        \label{NIPSDALfig:officehome7}
    \end{subfigure}
    \begin{subfigure}[b]{0.22\textwidth}\includegraphics[width=\textwidth]{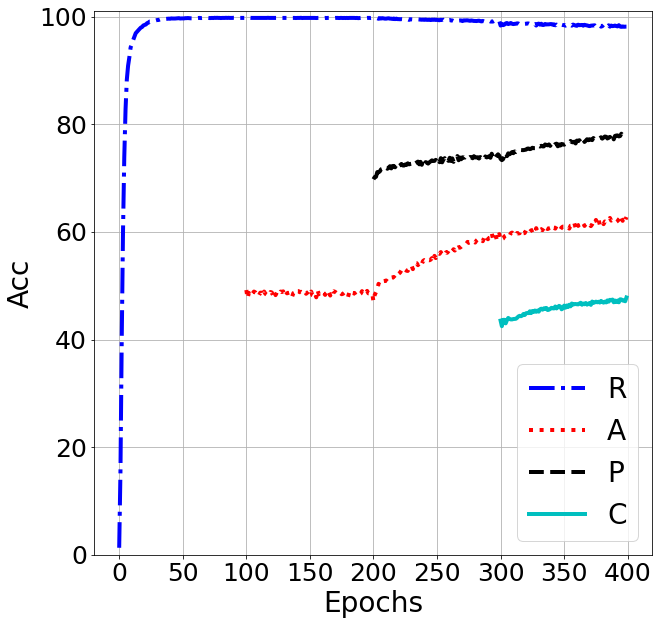}
           \centering
        \caption{$R \rightarrow A \rightarrow P \rightarrow  C  $}
        \label{NIPSDALfig:officehome8}
    \end{subfigure}\\    \centering
           \begin{subfigure}[b]{0.22\textwidth}\includegraphics[width=\textwidth]{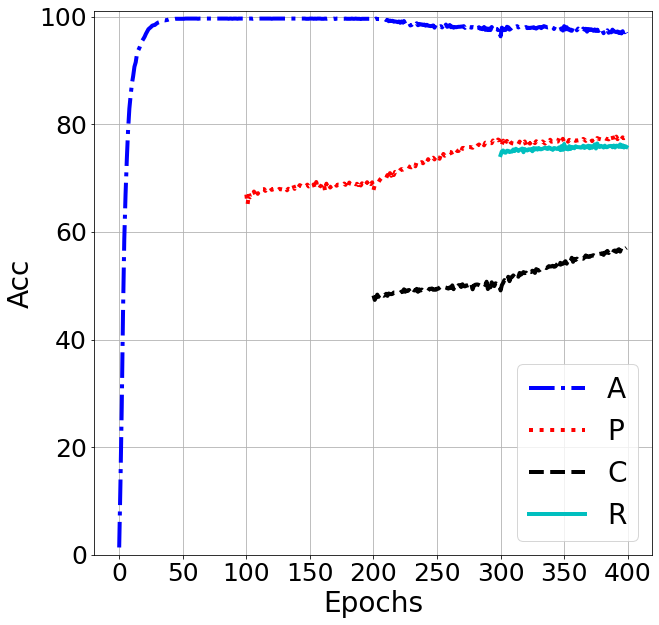}
           \centering
        \caption{$A \rightarrow P \rightarrow C \rightarrow  R  $}
        \label{NIPSDALfig:officehome9}
    \end{subfigure}
    \begin{subfigure}[b]{0.22\textwidth}\includegraphics[width=\textwidth]{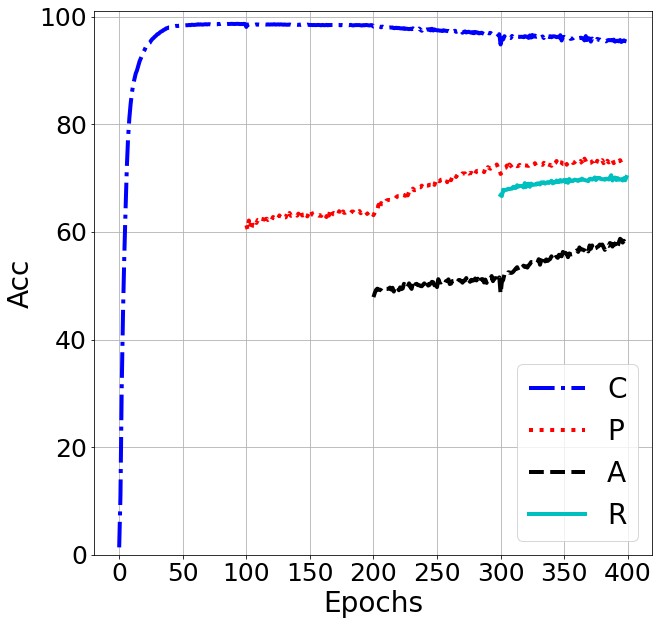}
           \centering
        \caption{$C \rightarrow P \rightarrow A \rightarrow  R  $}
        \label{NIPSDALfig:officehome10}
    \end{subfigure}
               \begin{subfigure}[b]{0.22\textwidth}\includegraphics[width=\textwidth]{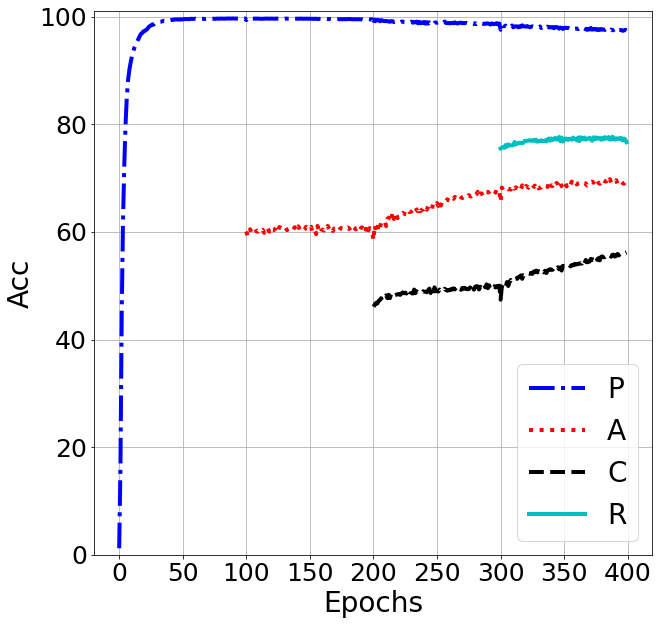}
           \centering
        \caption{$P \rightarrow C \rightarrow A \rightarrow  R  $}
        \label{NIPSDALfig:officehome11}
    \end{subfigure}
    \begin{subfigure}[b]{0.22\textwidth}\includegraphics[width=\textwidth]{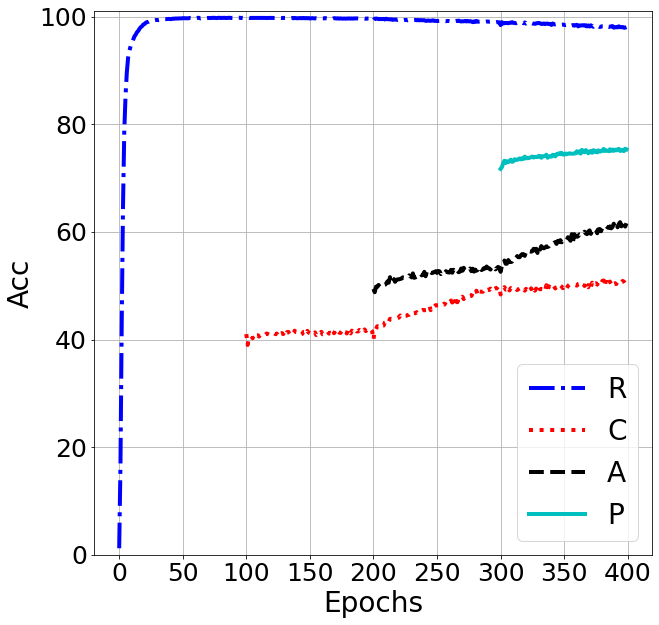}
           \centering
        \caption{$R \rightarrow C \rightarrow A \rightarrow  P  $}
        \label{NIPSDALfig:officehome12}
    \end{subfigure}\\    \centering
           \begin{subfigure}[b]{0.22\textwidth}\includegraphics[width=\textwidth]{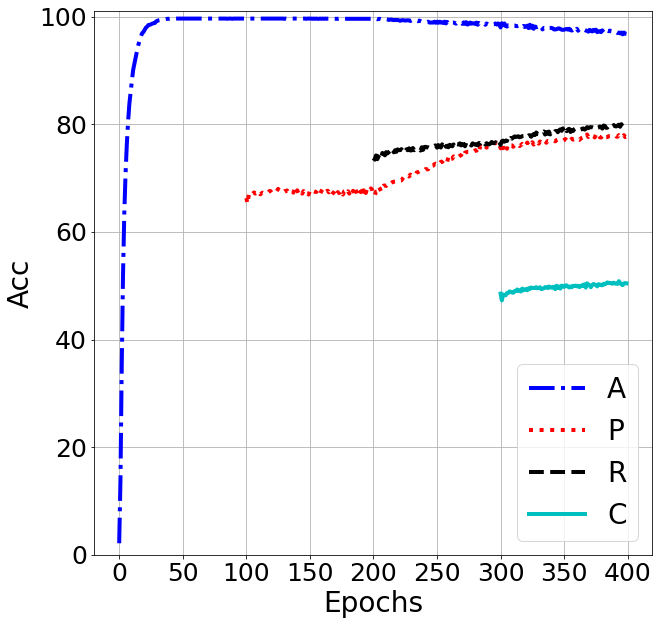}
           \centering
        \caption{$A \rightarrow P \rightarrow R \rightarrow  C  $}
        \label{NIPSDALfig:officehome13}
    \end{subfigure}
    \begin{subfigure}[b]{0.22\textwidth}\includegraphics[width=\textwidth]{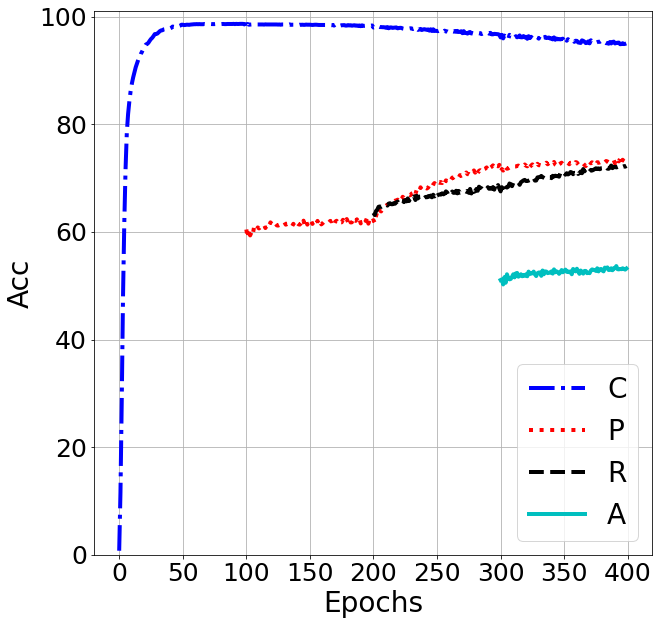}
           \centering
        \caption{$C \rightarrow P \rightarrow A \rightarrow  A  $}
        \label{NIPSDALfig:officehome14}
    \end{subfigure}
               \begin{subfigure}[b]{0.22\textwidth}\includegraphics[width=\textwidth]{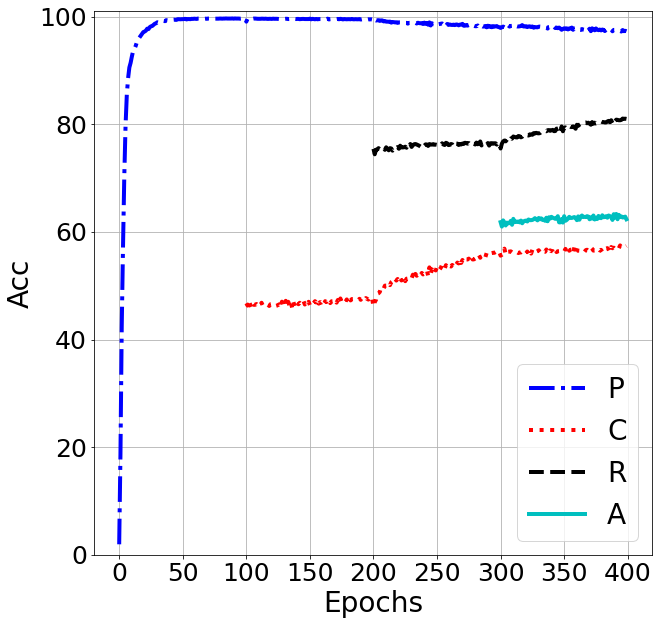}
           \centering
        \caption{$P \rightarrow C \rightarrow R \rightarrow  A  $}
        \label{NIPSDALfig:officehome15}
    \end{subfigure}
    \begin{subfigure}[b]{0.22\textwidth}\includegraphics[width=\textwidth]{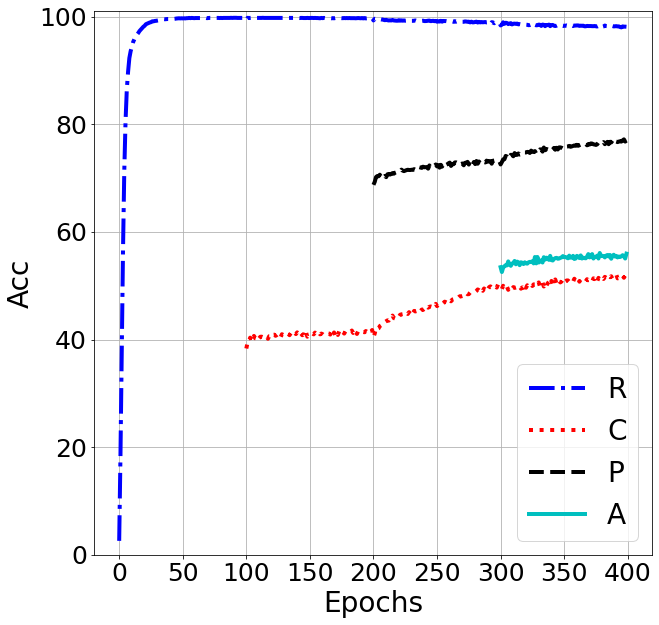}
           \centering
        \caption{$R \rightarrow C \rightarrow P \rightarrow  A  $}
        \label{NIPSDALfig:officehome16}
    \end{subfigure}\\    \centering
           \begin{subfigure}[b]{0.22\textwidth}\includegraphics[width=\textwidth]{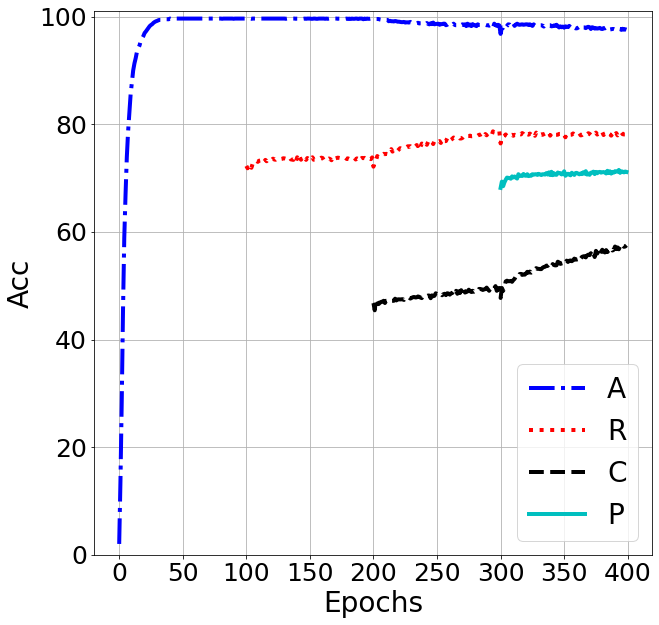}
           \centering
        \caption{$A \rightarrow R \rightarrow C \rightarrow  P  $}
        \label{NIPSDALfig:officehome17}
    \end{subfigure}
    \begin{subfigure}[b]{0.22\textwidth}\includegraphics[width=\textwidth]{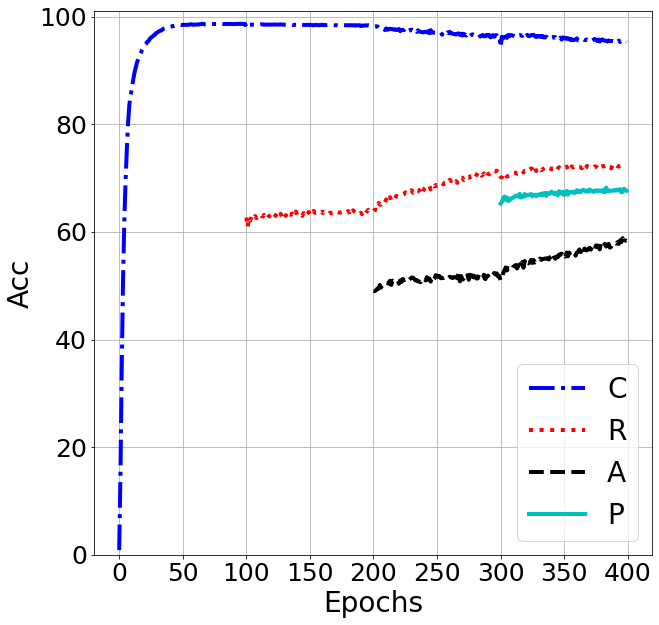}
           \centering
        \caption{$C \rightarrow R \rightarrow A \rightarrow  P  $}
        \label{NIPSDALfig:officehome18}
    \end{subfigure}
               \begin{subfigure}[b]{0.22\textwidth}\includegraphics[width=\textwidth]{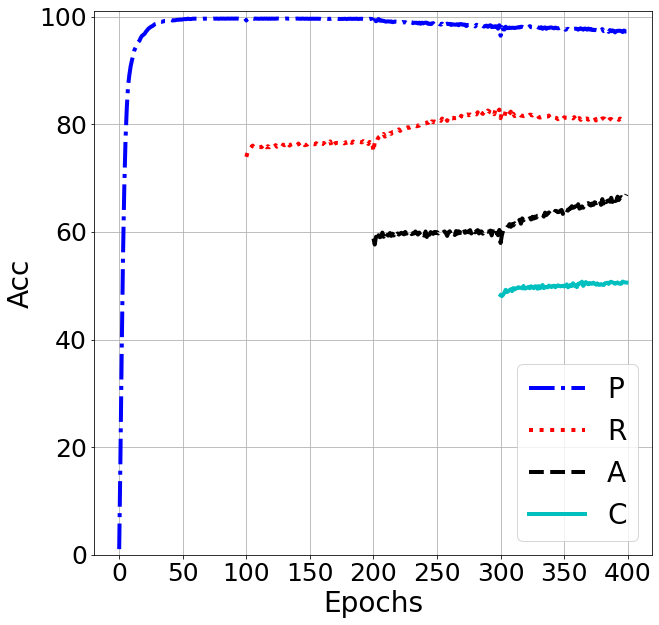}
           \centering
        \caption{$P \rightarrow R \rightarrow A \rightarrow  C  $}
        \label{NIPSDALfig:officehome19}
    \end{subfigure}
    \begin{subfigure}[b]{0.22\textwidth}\includegraphics[width=\textwidth]{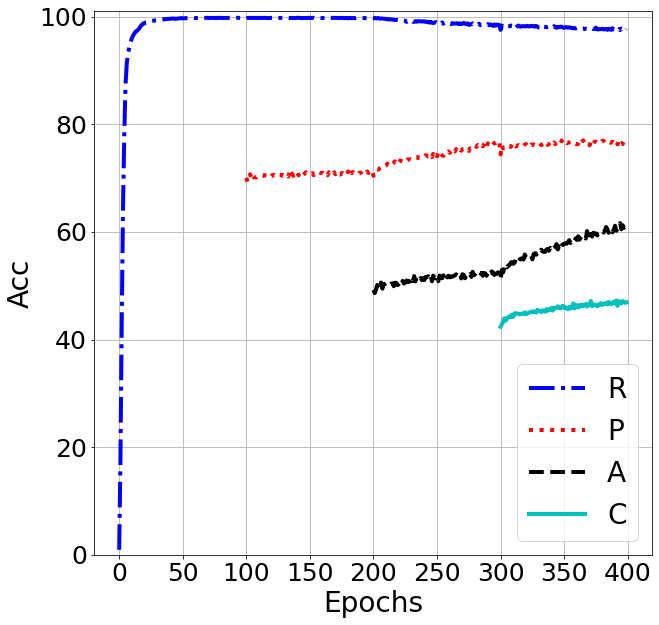}
           \centering
        \caption{$R \rightarrow P \rightarrow A \rightarrow  C  $}
        \label{NIPSDALfig:officehome20}
    \end{subfigure}\\    \centering
           \begin{subfigure}[b]{0.22\textwidth}\includegraphics[width=\textwidth]{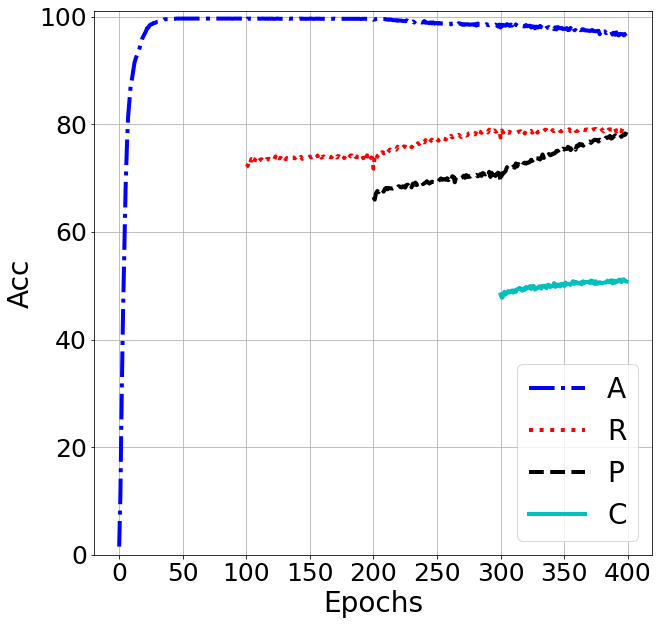}
           \centering
        \caption{$A \rightarrow R \rightarrow P \rightarrow  C  $}
        \label{NIPSDALfig:officehome21}
    \end{subfigure}
    \begin{subfigure}[b]{0.22\textwidth}\includegraphics[width=\textwidth]{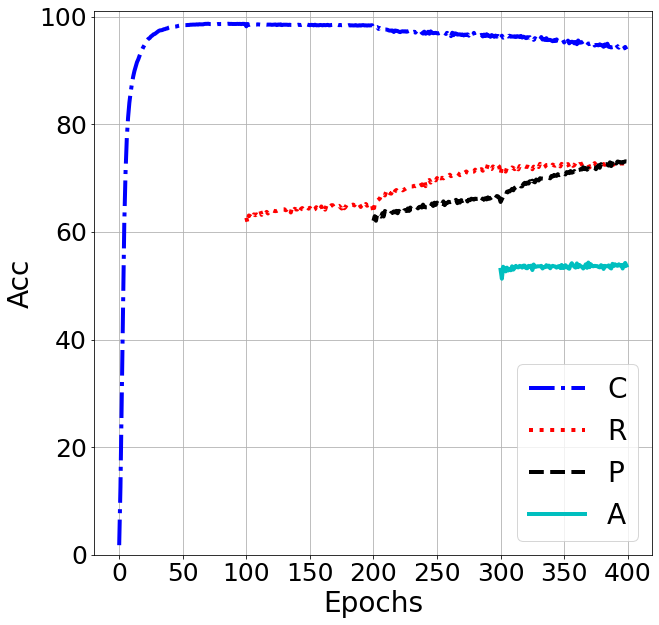}
           \centering
        \caption{$C \rightarrow R \rightarrow P \rightarrow  A  $}
        \label{NIPSDALfig:officehome22}
    \end{subfigure}
               \begin{subfigure}[b]{0.22\textwidth}\includegraphics[width=\textwidth]{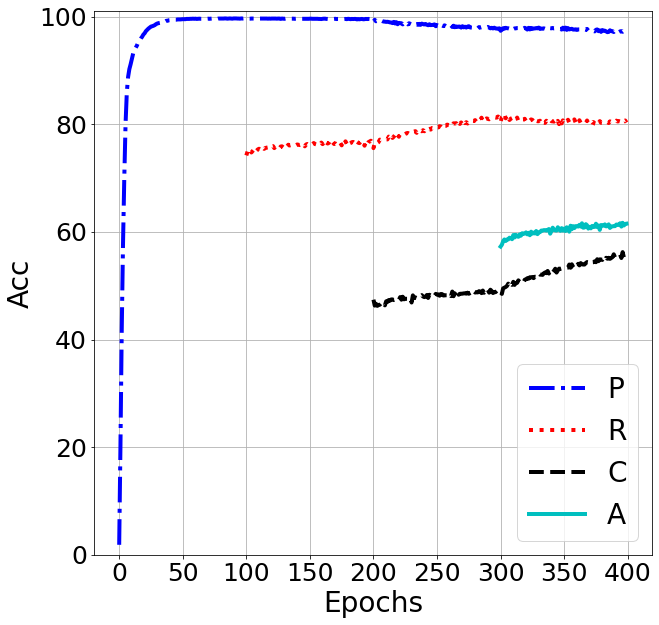}
           \centering
        \caption{$P \rightarrow R \rightarrow C \rightarrow  A  $}
        \label{NIPSDALfig:officehome23}
    \end{subfigure}
    \begin{subfigure}[b]{0.22\textwidth}\includegraphics[width=\textwidth]{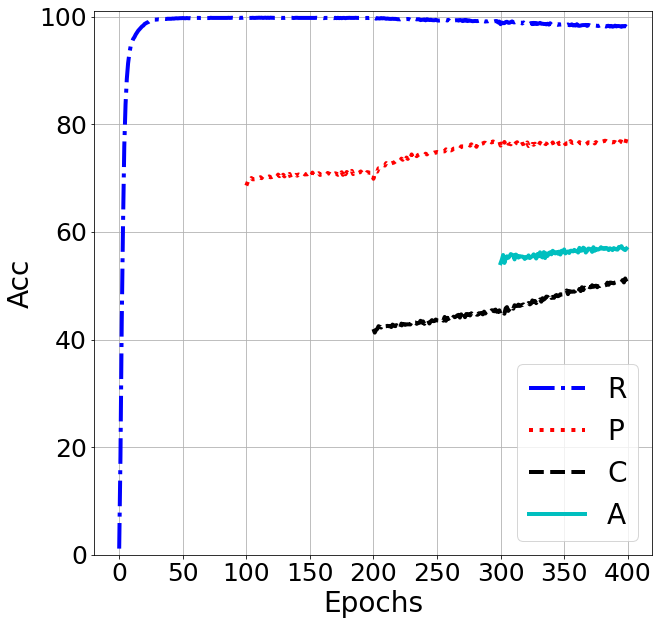}
           \centering
        \caption{$R \rightarrow P \rightarrow C \rightarrow  A  $}
        \label{NIPSDALfig:officehome24}
    \end{subfigure}
     \caption{Learning curves for sequential   tasks on  Office-Home  benchmark. (Best viewed in color).  }\label{NIPSDALfig:contrelating3}
\end{figure*}

The learning curves presented in Figure \ref{NIPSDALfig:contrelating3} for the 24 sequential UDA tasks of the Office-Home dataset demonstrate a similar trend to the previous findings, reaffirming the effectiveness of the LDAuCID algorithm in improving performance. In summary, the learning curves for the Office-Home dataset demonstrate a consistent overall trend, indicative of the effectiveness of the LDAuCID algorithm in improving performance and addressing the challenge of catastrophic forgetting.  Upon closer examination, several noteworthy distinctions come to light, which deserve careful consideration.
One notable observation is that the asymptotic performance improvements on the Office-Home dataset are relatively lower compared to those achieved through supervised learning. For instance, when the domain $A$ is learned as the second task (as depicted in Figure \ref{NIPSDALfig:officehome2}), compared to being learned as the last task (as shown in Figure \ref{NIPSDALfig:officehome1}), we observe a performance gap of approximately 20\%. This discrepancy in performance can be attributed to the presence of larger domain gaps between the domains in this dataset. The substantial domain gap poses a greater challenge for UDA, resulting in comparatively lower performance gains. However,  this observation is not specific to our   algorithm. When compared with other UDA methods in the subsequent section, our performance remains competitive within the UDA setting. This observation underscores the sensitivity of model performance to the order in which domains are encountered.
Another notable distinction is that in all 24 tasks is the successful mitigation of forgetting effects, accompanied by positive backward knowledge transfer across the past learned tasks. This finding highlights the algorithm's capacity to retain previously acquired knowledge while adapting the model to new domains.

In conclusion, the LDAuCID algorithm proves highly effective in mitigating forgetting effects and improving generalizability within the Office-Home benchmark dataset. While the performance gains in UDA tasks may be comparatively lower due to larger domain gaps, the algorithm remains competitive when compared to other UDA methods. 

\begin{figure*} 
\centering
       \begin{subfigure}[b]{0.22\textwidth}\includegraphics[width=\textwidth]{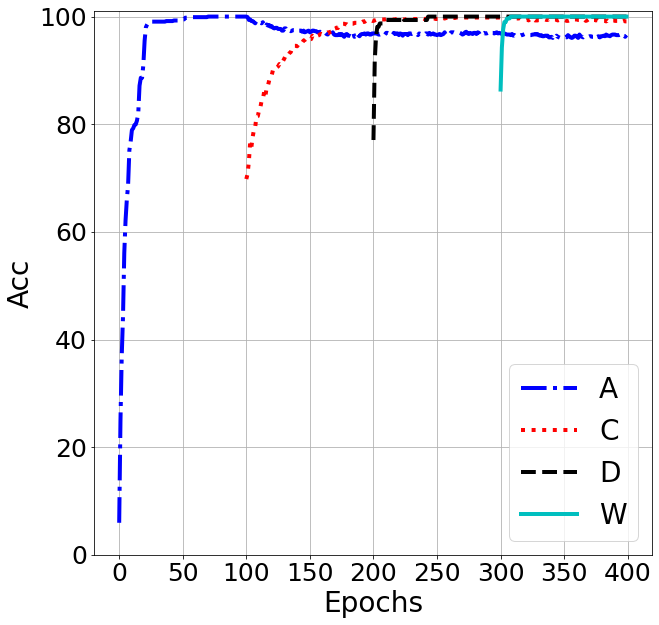}
           \centering
        \caption{$A \rightarrow C \rightarrow D \rightarrow  W  $}
        \label{NIPSDALfig:OfficeCaltech1}
    \end{subfigure}
       \begin{subfigure}[b]{0.22\textwidth}\includegraphics[width=\textwidth]{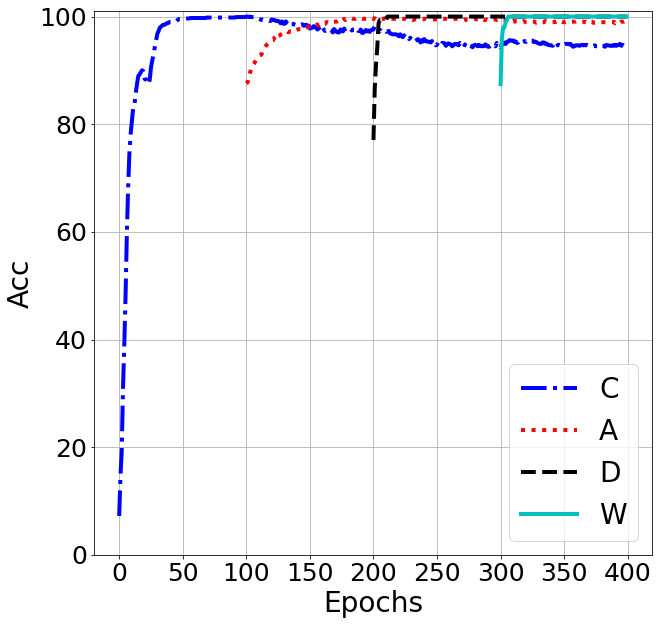}
              \centering
        \caption{$C \rightarrow A \rightarrow D \rightarrow  W  $}
        \label{NIPSDALfig:OfficeCaltech2}
    \end{subfigure}
           \begin{subfigure}[b]{0.22\textwidth}\includegraphics[width=\textwidth]{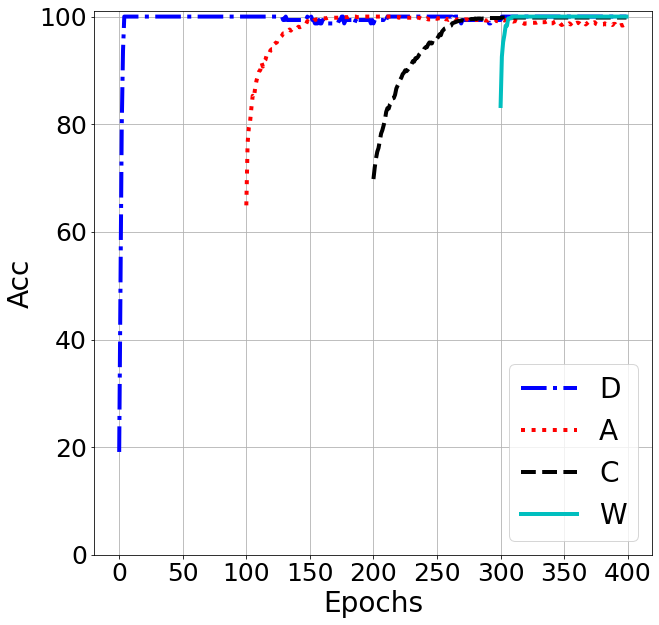}
           \centering
        \caption{$D \rightarrow A \rightarrow C \rightarrow  W  $}
        \label{NIPSDALfig:OfficeCaltech3}
    \end{subfigure}
       \begin{subfigure}[b]{0.22\textwidth}\includegraphics[width=\textwidth]{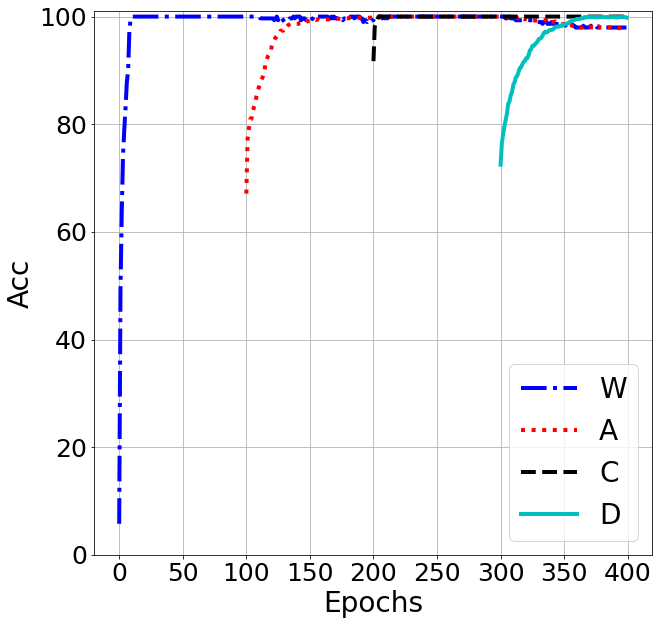}
              \centering
        \caption{$W \rightarrow A \rightarrow C \rightarrow  D  $}
        \label{NIPSDALfig:OfficeCaltech4}
    \end{subfigure}\\
    \centering
       \begin{subfigure}[b]{0.22\textwidth}\includegraphics[width=\textwidth]{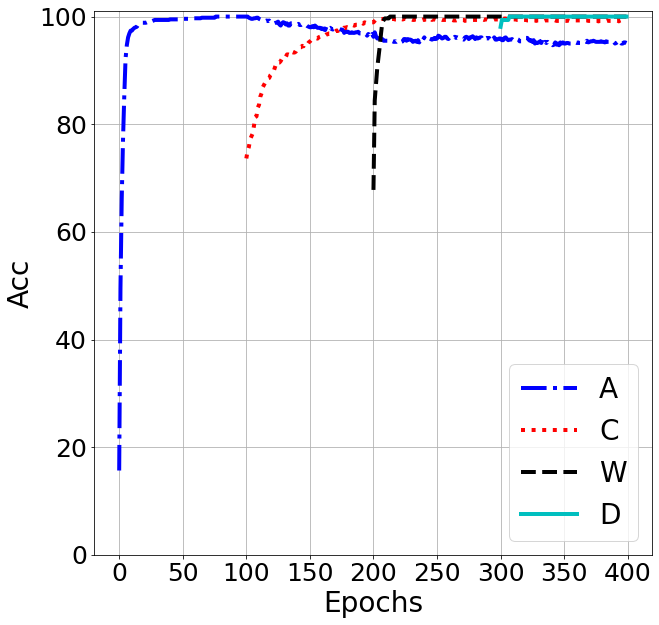}
           \centering
        \caption{$A \rightarrow C \rightarrow W \rightarrow  D  $}
        \label{NIPSDALfig:OfficeCaltech5}
    \end{subfigure}
       \begin{subfigure}[b]{0.22\textwidth}\includegraphics[width=\textwidth]{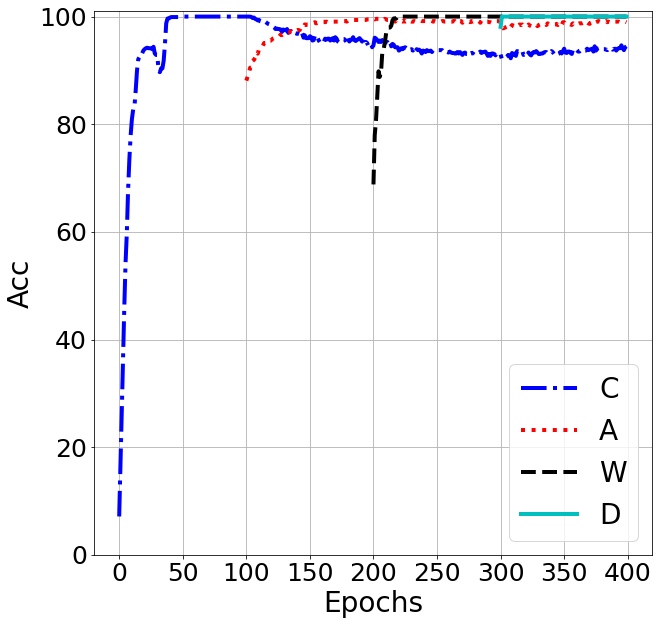}
              \centering
        \caption{$C \rightarrow A \rightarrow W \rightarrow  D  $}
        \label{NIPSDALfig:OfficeCaltech6}
    \end{subfigure}
           \begin{subfigure}[b]{0.22\textwidth}\includegraphics[width=\textwidth]{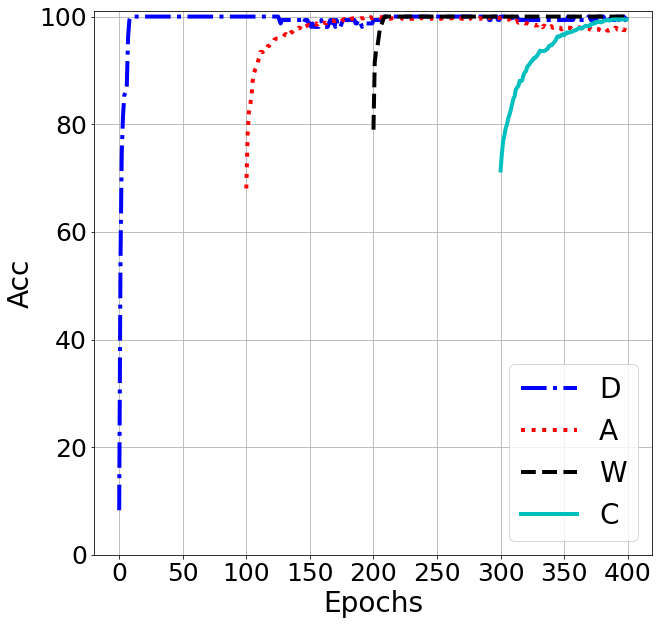}
           \centering
        \caption{$D \rightarrow A \rightarrow W \rightarrow  C  $}
        \label{NIPSDALfig:OfficeCaltech7}
    \end{subfigure}
       \begin{subfigure}[b]{0.22\textwidth}\includegraphics[width=\textwidth]{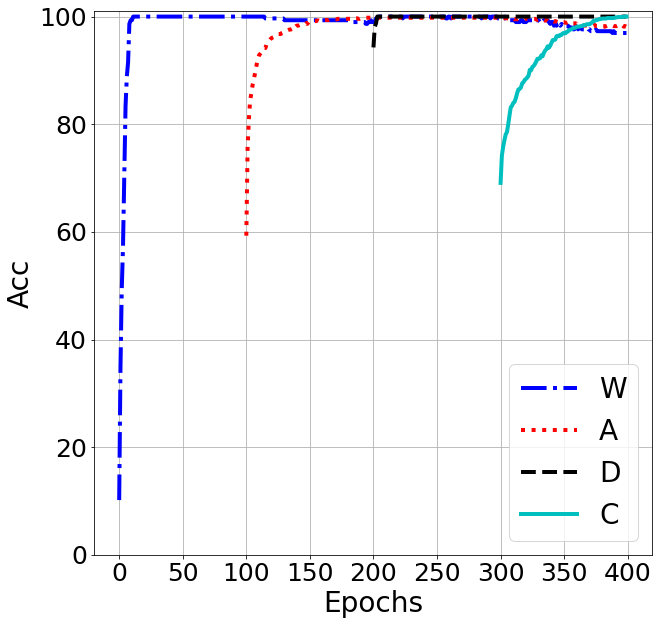}
              \centering
        \caption{$W \rightarrow A \rightarrow D \rightarrow  C  $}
        \label{NIPSDALfig:OfficeCaltech8}
    \end{subfigure}\\
    \centering
       \begin{subfigure}[b]{0.22\textwidth}\includegraphics[width=\textwidth]{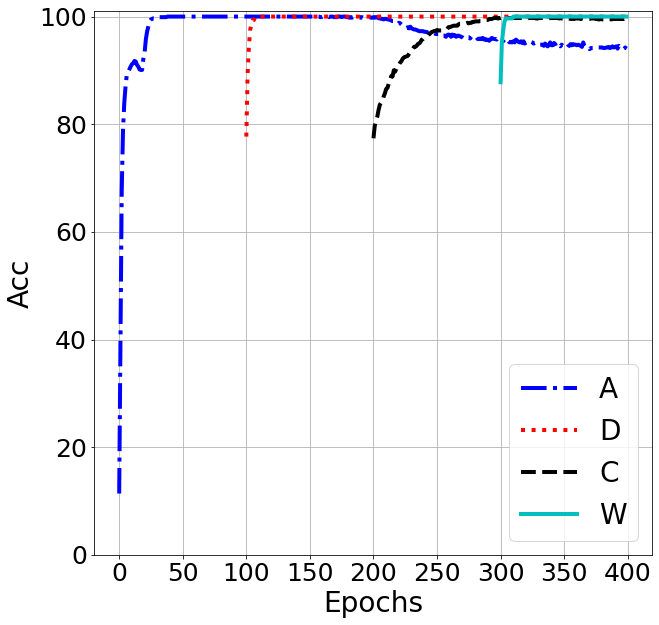}
           \centering
        \caption{$A \rightarrow D \rightarrow C \rightarrow  W  $}
        \label{NIPSDALfig:OfficeCaltech9}
    \end{subfigure}
       \begin{subfigure}[b]{0.22\textwidth}\includegraphics[width=\textwidth]{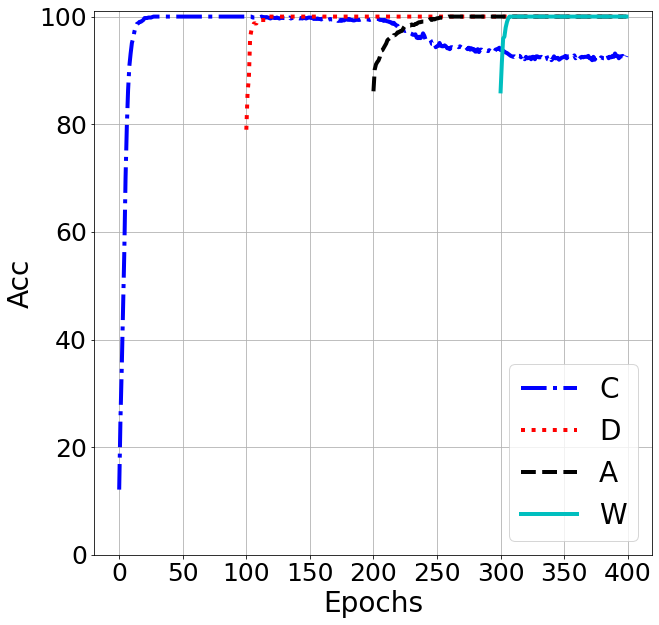}
              \centering
        \caption{$C \rightarrow D \rightarrow A \rightarrow  W  $}
        \label{NIPSDALfig:OfficeCaltech10}
    \end{subfigure}
           \begin{subfigure}[b]{0.22\textwidth}\includegraphics[width=\textwidth]{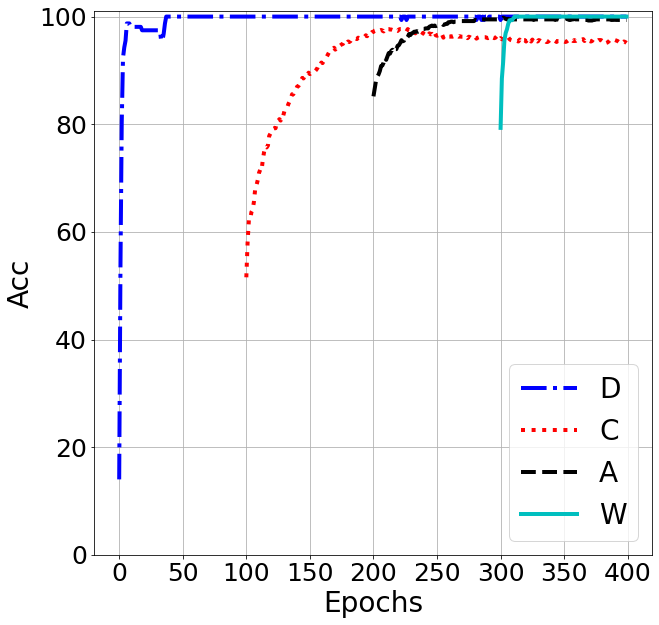}
           \centering
        \caption{$D \rightarrow C \rightarrow A \rightarrow  W  $}
        \label{NIPSDALfig:OfficeCaltech11}
    \end{subfigure}
       \begin{subfigure}[b]{0.22\textwidth}\includegraphics[width=\textwidth]{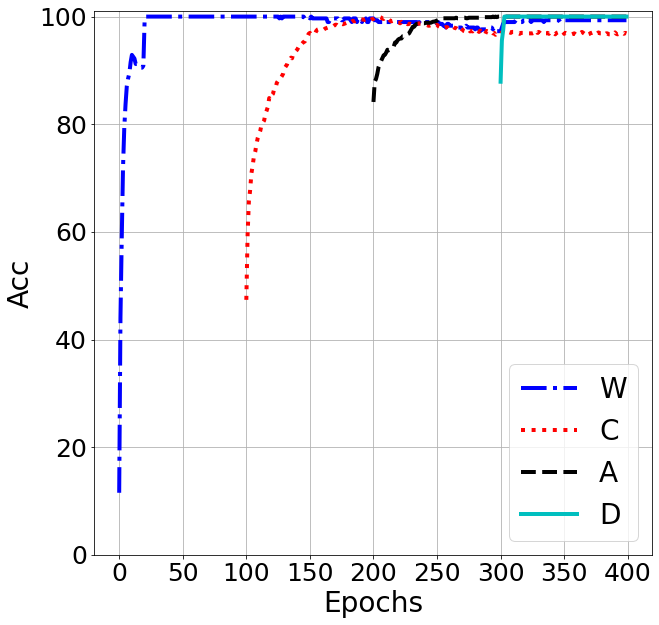}
              \centering
        \caption{$W \rightarrow C \rightarrow A \rightarrow  D  $}
        \label{NIPSDALfig:OfficeCaltech12}
    \end{subfigure}\\
    \centering
       \begin{subfigure}[b]{0.22\textwidth}\includegraphics[width=\textwidth]{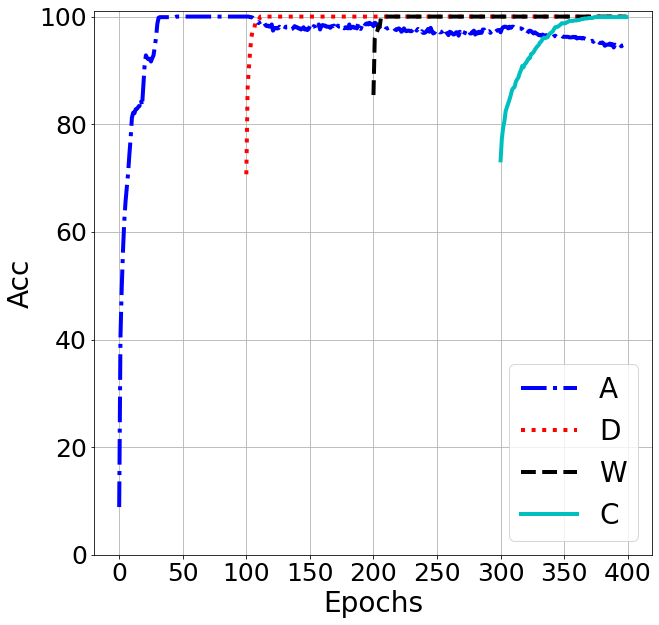}
           \centering
        \caption{$A \rightarrow D \rightarrow W \rightarrow  C  $}
        \label{NIPSDALfig:OfficeCaltech13}
    \end{subfigure}
       \begin{subfigure}[b]{0.22\textwidth}\includegraphics[width=\textwidth]{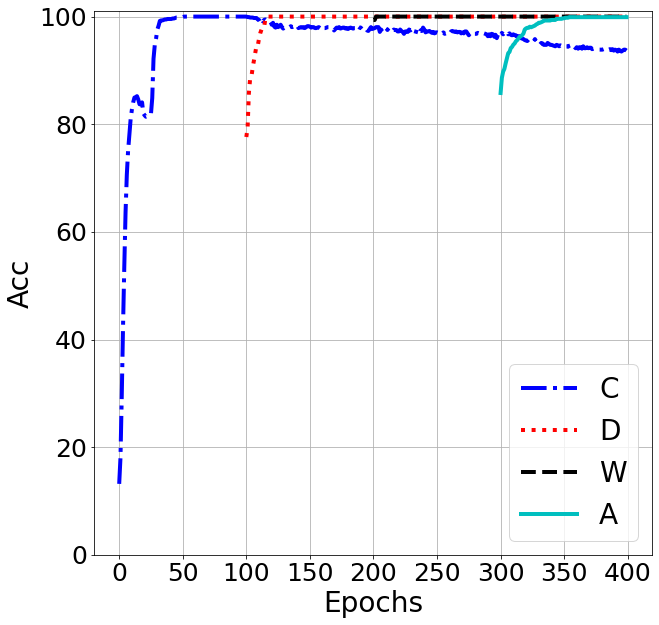}
              \centering
        \caption{$C \rightarrow D \rightarrow W \rightarrow  A  $}
        \label{NIPSDALfig:OfficeCaltech14}
    \end{subfigure}
           \begin{subfigure}[b]{0.22\textwidth}\includegraphics[width=\textwidth]{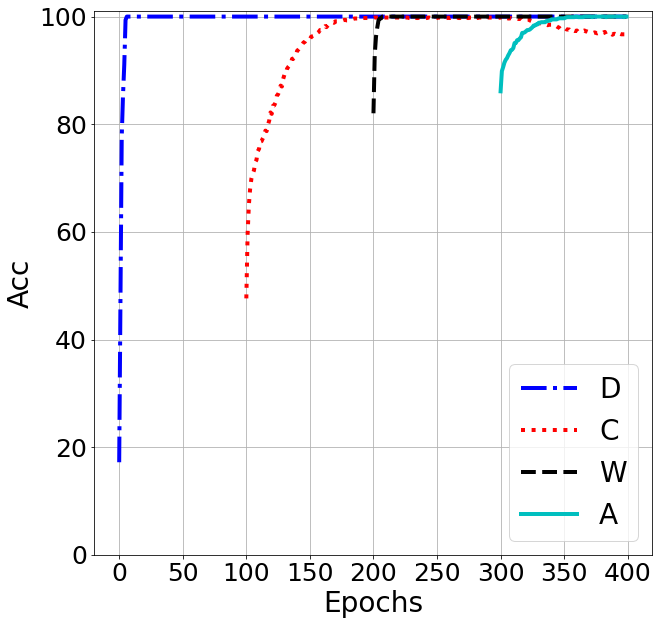}
           \centering
        \caption{$D \rightarrow C \rightarrow W \rightarrow  A  $}
        \label{NIPSDALfig:OfficeCaltech15}
    \end{subfigure}
       \begin{subfigure}[b]{0.22\textwidth}\includegraphics[width=\textwidth]{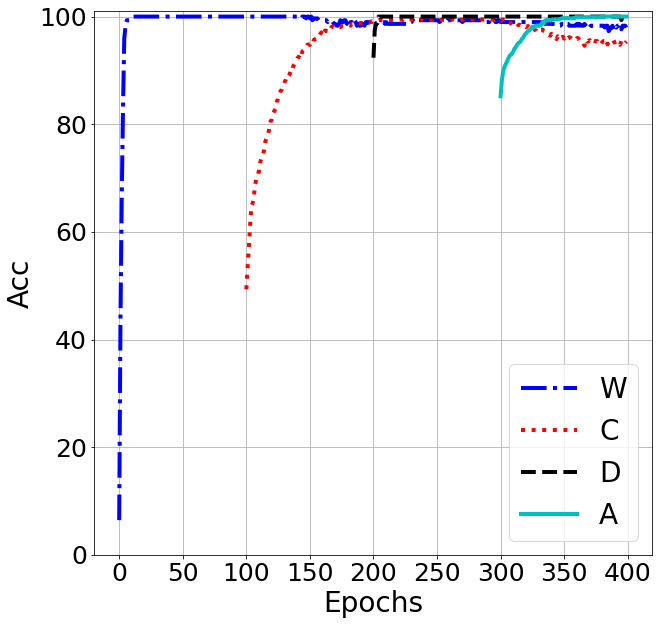}
              \centering
        \caption{$W \rightarrow C \rightarrow D \rightarrow  A  $}
        \label{NIPSDALfig:OfficeCaltech16}
    \end{subfigure}\\
    \centering
       \begin{subfigure}[b]{0.22\textwidth}\includegraphics[width=\textwidth]{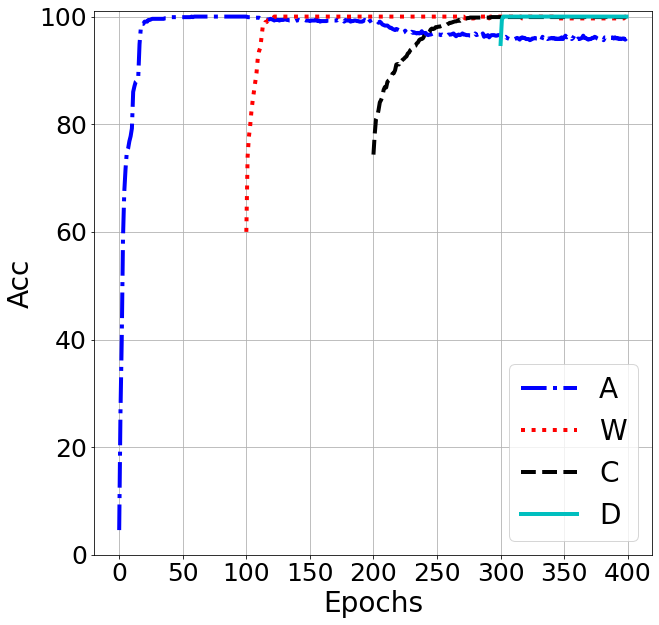}
           \centering
        \caption{$A \rightarrow W \rightarrow C \rightarrow  D  $}
        \label{NIPSDALfig:OfficeCaltech17}
    \end{subfigure}
       \begin{subfigure}[b]{0.22\textwidth}\includegraphics[width=\textwidth]{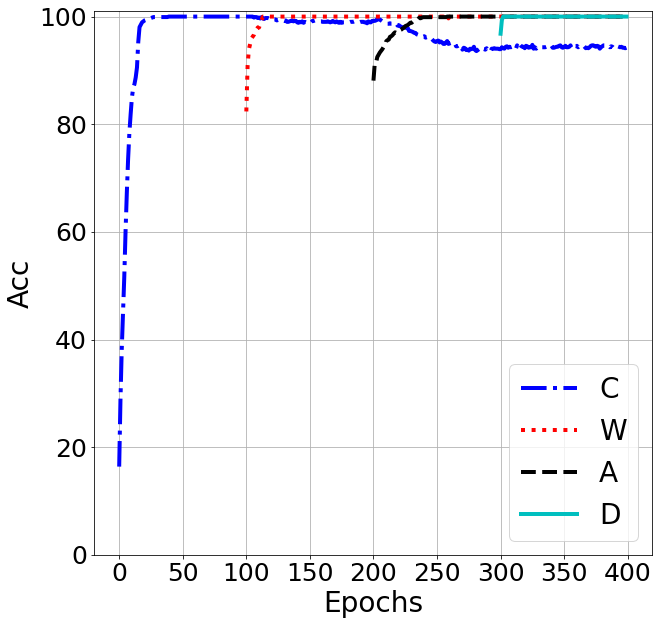}
              \centering
        \caption{$C \rightarrow W \rightarrow A \rightarrow  D  $}
        \label{NIPSDALfig:OfficeCaltech18}
    \end{subfigure}
           \begin{subfigure}[b]{0.22\textwidth}\includegraphics[width=\textwidth]{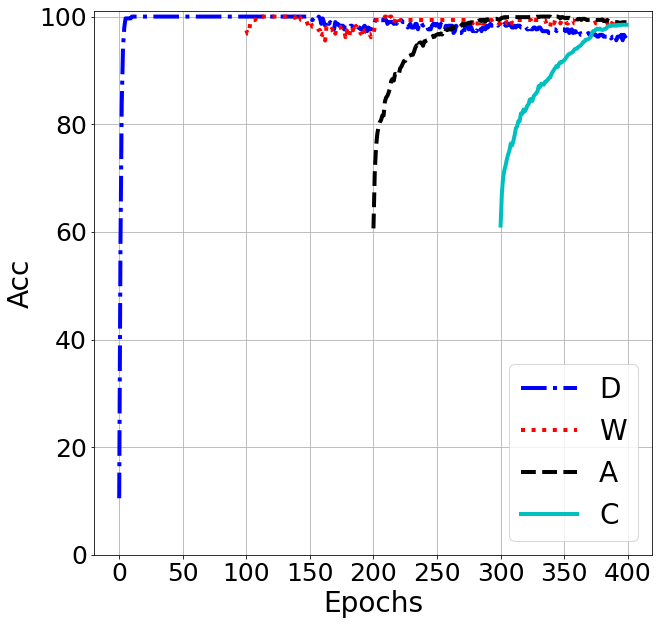}
           \centering
        \caption{$D \rightarrow W \rightarrow A \rightarrow  C  $}
        \label{NIPSDALfig:OfficeCaltech19}
    \end{subfigure}
       \begin{subfigure}[b]{0.22\textwidth}\includegraphics[width=\textwidth]{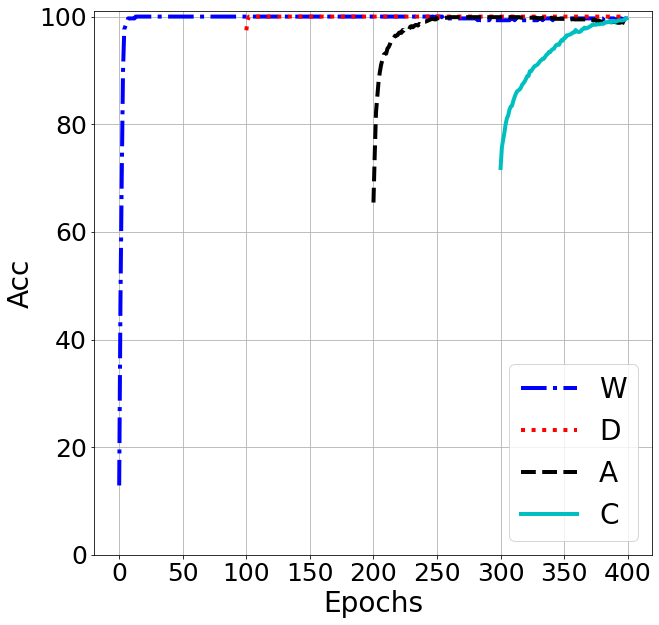}
              \centering
        \caption{$W \rightarrow D \rightarrow A \rightarrow  C  $}
        \label{NIPSDALfig:OfficeCaltech20}
    \end{subfigure}\\
    \centering
       \begin{subfigure}[b]{0.22\textwidth}\includegraphics[width=\textwidth]{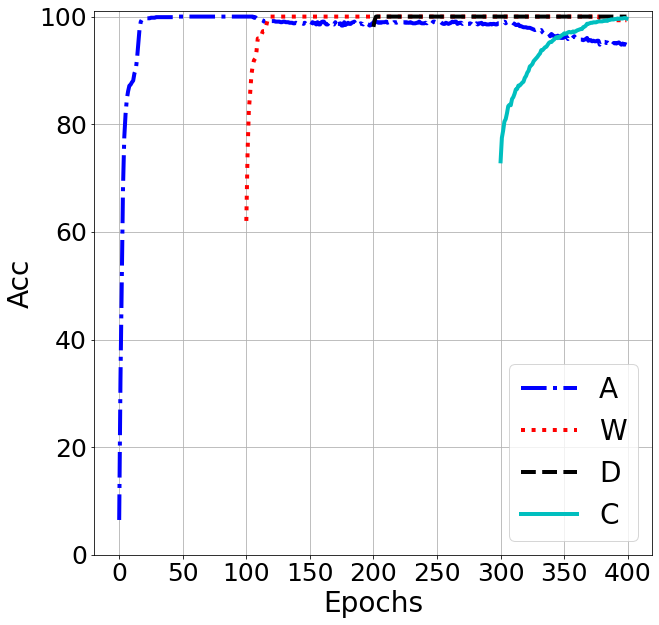}
           \centering
        \caption{$A \rightarrow W \rightarrow D \rightarrow  C  $}
        \label{NIPSDALfig:OfficeCaltech21}
    \end{subfigure}
       \begin{subfigure}[b]{0.22\textwidth}\includegraphics[width=\textwidth]{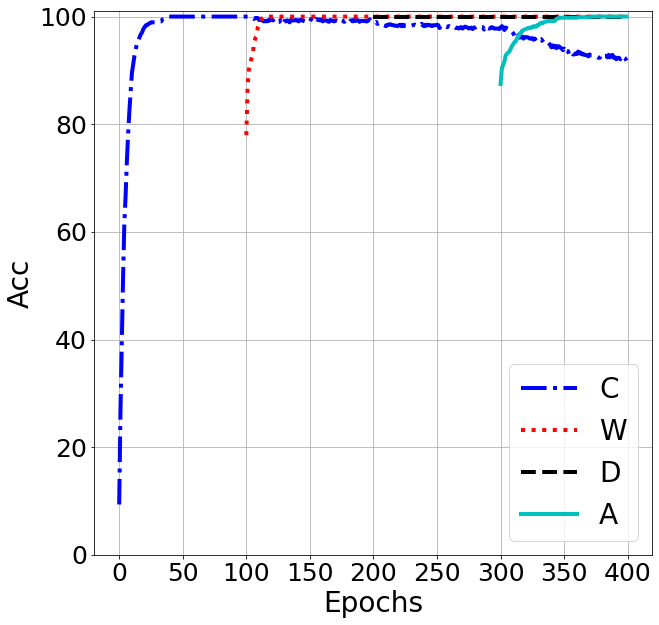}
              \centering
        \caption{$C \rightarrow W \rightarrow D \rightarrow  A  $}
        \label{NIPSDALfig:OfficeCaltech22}
    \end{subfigure}
           \begin{subfigure}[b]{0.22\textwidth}\includegraphics[width=\textwidth]{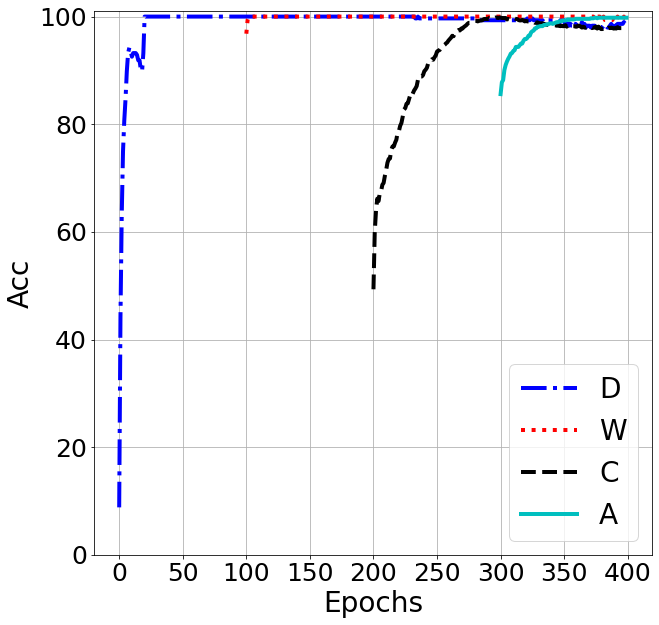}
           \centering
        \caption{$D \rightarrow W \rightarrow C \rightarrow  A  $}
        \label{NIPSDALfig:OfficeCaltech23}
    \end{subfigure}
       \begin{subfigure}[b]{0.22\textwidth}\includegraphics[width=\textwidth]{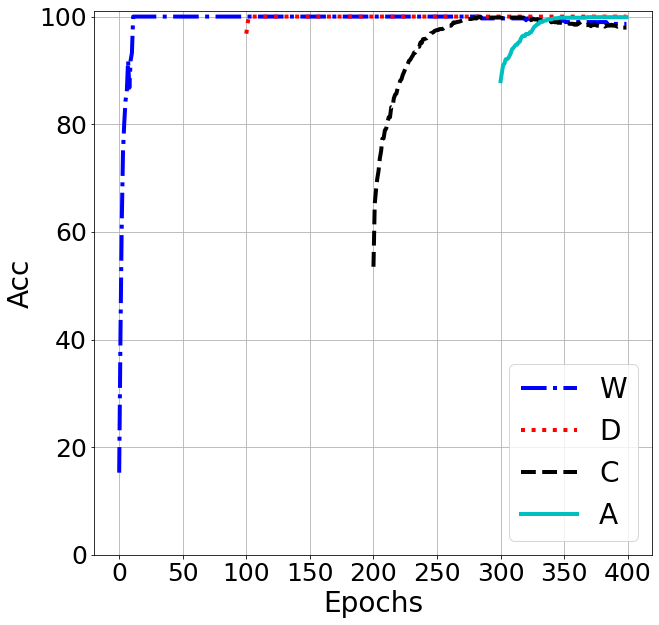}
              \centering
        \caption{$W \rightarrow D \rightarrow C \rightarrow  A  $}
        \label{NIPSDALfig:OfficeCaltech24}
    \end{subfigure}
     \caption{Learning curves for sequential UDA tasks on   Office-Caltech benchmark. (Best viewed in color).  }\label{NIPSDALfig:contrelating4}
\end{figure*}

Figure \ref{NIPSDALfig:contrelating4} illustrates the learning curves for the Office-Caltech benchmark. Remarkably, our algorithm demonstrates great performance, particularly by minimal catastrophic forgetting and relatively high final performance across most tasks.
One key factor contributing to the successful performance of our algorithm in this benchmark is the simplicity and similarities between the domains, i.e., the domain  gap between the tasks is smaller compared to the Office-Home benchmarks. The domain shift in this dataset is relatively smaller compared to the other benchmarks, which allows for easier knowledge transfer and adaptation. As a result, the model can effectively retain the learned information from previous tasks and maintain stable performance without suffering from significant forgetting.
This observation aligns with our theoretical analysis, which emphasizes the importance of the tightness of the upperbound in relation to the expected target error. In this case, the smaller domain shift leads to a more favorable upperbound, enabling the model to leverage the shared knowledge and generalize well to new tasks. Hence, the performance of our algorithm can also be influenced by factors and properties that are beyond our control or design choices. In the case of the Office-Caltech benchmark, the inherent characteristics of the dataset, such as the simplicity of domain similarities, contribute to the superior performance of LDAuCID. These external factors highlight the complexity of UDA tasks and the impact of dataset properties on algorithmic performance.

\subsubsection{Comparative Results in UDA settings}

Table~\ref{table:tabDA1} provides a comprehensive overview of the comparative results for six bi-domain digit recognition tasks. The table includes performance metrics for each method, with the Source-Only approach serving as the baseline. The Source-Only performance represents the model's performance on the target domain without any adaptation. Improvements over the Source-Only perofrmnace indicates the effectiveness of a UDA algorithm.
Interestingly, despite the classic UDA methods utilizing the full source dataset for joint-training, we observe that LDAuCID outperforms the UDA methods in two of the tasks and remains competitive in the remaining tasks. This result showcases the effectiveness of our proposed algorithm in improving domain adaptation performance even in the classical setting.
Additionally, it is worth noting that the ETD method consistently demonstrates a high performance across the tasks. This outcome is not surprising, as ETD shares similarities with our approach. ETD incorporates a variation of the WD distance for enhancing the classic UDA joint-training scheme. The superior performance of ETD can be attributed to its ability to effectively leverage the source data directly, in contrast to our method, and optimize the model for domain adaptation.
The comparative results presented in Table~\ref{table:tabDA1} highlight the strengths of LDAuCID in the context of bi-domain digit recognition tasks. Despite the advantages of classic UDA methods that leverage the full source dataset, our algorithm surpasses their performance in multiple tasks and remains competitive in others. Furthermore, the prominence of ETD reaffirms the effectiveness of incorporating WD distance in the UDA framework.

  \begin{table*}[t!]
 \setlength{\tabcolsep}{1pt}
 \centering 
{\footnotesize
\begin{tabular}{lc|ccc|c|ccc}   
\multicolumn{2}{c}{Method}    & $\mathcal{M}\rightarrow\mathcal{U}$ & $\mathcal{U}\rightarrow\mathcal{M}$ & $\mathcal{S}\rightarrow\mathcal{M}$ &Method & $\mathcal{M}\rightarrow\mathcal{U}$ & $\mathcal{U}\rightarrow\mathcal{M}$ & $\mathcal{S}\rightarrow\mathcal{M}$\\
\hline
\multicolumn{2}{c|}{GtA~ \cite{sankaranarayanan2018generate}}& 92.8  $\pm$  0.9	&	90.8  $\pm$  1.3	&	92.4  $\pm$  0.9 &CDAN~ \cite{long2018conditional} &93.9 &96.9& 88.5 \\
\multicolumn{2}{c|}{CoGAN~ \cite{liu2016coupled}}& 91.2  $\pm$  0.8&89.1  $\pm$  0.8&-& SHOT~ \cite{liang2020we} &89.6$\pm$5.0   &96.8$\pm$0.4  & 91.9$\pm$0.4 \\ 
\multicolumn{2}{c|}{ADDA~ \cite{tzeng2017adversarial}}& 89.4  $\pm$  0.2&90.1  $\pm$  0.8&76.0  $\pm$  1.8& CyCADA~ \cite{hoffman2017cycada}&   95.6   $\pm$  0.2  & 96.5  $\pm$  0.1 & 90.4  $\pm$  0.4  \\  
\multicolumn{2}{c|}{RevGrad~ \cite{ganin2014unsupervised}}&	 77.1  $\pm$  1.8 	&	73.0  $\pm$  2.0 	&	73.9  &JDDA~ \cite{chen2019joint} & -& 97.0 $\pm$0.2 &  93.1  $\pm$0.2   \\ 
\multicolumn{2}{c|}{DRCN~ \cite{ghifary2016deep}}&	 91.8  $\pm$  0.1 	&	73.7  $\pm$  0.4 	&	82.0  $\pm$  0.2 &  OPDA~ \cite{courty2017optimal}& 70.0 & 60.2 &-   \\
\multicolumn{2}{c|}{ETD~ \cite{li2020enhanced}}& 96.4$\pm$ 0.3 & 96.3$\pm$ 0.1&\textbf{97.9}$\pm$ 0.4&
MML~ \cite{seguy2017large}& 77.9  & 60.5  &62.9  \\ 
\hline
\multicolumn{2}{c|}{Source Only}&	 90.1$\pm$2.6	&	80.2$\pm$5.7	&	67.3$\pm$2.6   & LDAuCID  &   \textbf{96.8}  $\pm$  0.2	&	\textbf{98.4}  $\pm$  0.1	&	91.4  $\pm$  2.2 \\
\end{tabular}}
\caption{ Classification accuracy for UDA tasks between MNIST, USPS, and SVHN    datasets.    }
\label{table:tabDA1}
 \end{table*}

The performance results for the ImageCLEF-DA dataset tasks are summarized in Table~\ref{table:tabDA3}. Notably, LDAuCID has demonstrated a substantial improvement on this particular dataset. The ImageCLEF-DA dataset is characterized by its balanced nature, both in terms of the number of data points per domain and per class. This balanced distribution plays a significant role in the competitive performance of our method, as we rely on the empirical versions of the distributions for domain alignment. The balanced nature of the dataset ensures that the empirical distribution is a less biased approximation of the true distribution.
Based on our observations, we can conclude that having a balanced dataset across the domains can greatly enhance the performance of LDAuCID. The unbiased estimation of the empirical distributions allows for more accurate alignment between the domains, resulting in improved performance.
Therefore, the results obtained from the ImageCLEF-DA dataset reinforce the importance of dataset balance in the effectiveness of our method.

  \begin{table*}[t!]
  \setlength{\tabcolsep}{1pt}
 \centering 
{\footnotesize
\begin{tabular}{lc|cccccc|c}   
\multicolumn{2}{c}{Method}    & $\mathcal{I}\rightarrow\mathcal{P}$ & $\mathcal{P}\rightarrow\mathcal{I}$ & $\mathcal{I}\rightarrow\mathcal{C}$ &$\mathcal{C}\rightarrow\mathcal{I}$ &$\mathcal{C}\rightarrow\mathcal{P}$ &$\mathcal{P}\rightarrow\mathcal{C}$& Average \\
\hline
\multicolumn{2}{c|}{Source Only~ \cite{he2016deep}}& 74.8  $\pm$  0.3& 83.9  $\pm$  0.1& 91.5  $\pm$  0.3 &78.0  $\pm$  0.2 &65.5  $\pm$  0.3& 91.2  $\pm$  0.3&80.8  \\
\multicolumn{2}{c|}{DANN~ \cite{ganin2016domain}}&   82.0  $\pm$  0.4& 96.9  $\pm$  0.2& 99.1  $\pm$  0.1& 79.7  $\pm$  0.4& 68.2  $\pm$  0.4 &67.4  $\pm$  0.5 &82.2 \\ 
\multicolumn{2}{c|}{MADA~ \cite{pei2018multi}}& 75.0 $\pm$ 0.3& 87.9 $\pm$ 0.2& 96.0 $\pm$ 0.3& 88.8 $\pm$ 0.3& 75.2 $\pm$ 0.2& 92.2 $\pm$ 0.3   & 85.9 \\
\multicolumn{2}{c|}{CDAN~ \cite{long2018conditional} }&76.7 $\pm$ 0.3& 90.6 $\pm$ 0.3& 97.0 $\pm$ 0.4 &90.5 $\pm$ 0.4& 74.5 $\pm$ 0.3 &93.5 $\pm$ 0.4 &87.1\\  
\multicolumn{2}{c|}{DAN~ \cite{long2015learning}}& 74.5 $\pm$ 0.4 &82.2 $\pm$ 0.2 &92.8 $\pm$ 0.2& 86.3 $\pm$ 0.4& 69.2 $\pm$ 0.4& 89.8 $\pm$ 0.4&82.4\\ 
\multicolumn{2}{c|}{RevGrad~ \cite{ganin2014unsupervised}} &75.0 $\pm$ 0.6 &86.0 $\pm$ 0.3& 96.2 $\pm$ 0.4 &87.0 $\pm$ 0.5& 74.3 $\pm$ 0.5& 91.5 $\pm$ 0.6&85.0 \\
\multicolumn{2}{c|}{JAN~ \cite{long2017deep}}&  76.8 $\pm$ 0.4&  88.0 $\pm$ 0.2&  94.7 $\pm$ 0.2 & 89.5 $\pm$ 0.3&  74.2 $\pm$ 0.3 & 91.7 $\pm$ 0.3&85.7\\ %
\multicolumn{2}{c|}{ETD~ \cite{li2020enhanced}}&  81.0 & 91.7 & 97.9 & 93.3 & 79.5 & 95.0&  89.7 \\ 
\hline
\multicolumn{2}{c|}{LDAuCID}&	 		\textbf{87.8}  $\pm$  1.4	&	\textbf{99.1}  $\pm$  0.2 & \textbf{100}  $\pm$  0.0   & \textbf{99.8}  $\pm$  0.0 & \textbf{88.8}  $\pm$  1.0 & \textbf{99.5}  $\pm$  0.3     & \textbf{95.8}
\\
\end{tabular}}
\caption{ Classification accuracy for UDA tasks for  ImageCLEF-DA dataset. }
\label{table:tabDA3}
 \end{table*}
   
The performance results for the Office-Home dataset are summarized in Table~\ref{table:tabDA4}. CDAN exhibits the highest average performance among the methods, but LDAuCID remains competitive and even outperforms CDAN in four of the tasks. As previously discussed, the domains in this dataset exhibit a larger gap compared to the other datasets we considered. This observation is supported by Figure~\ref{NIPSDALfig:contrelating3}, where we can see a lower jumpstart performance value.
The larger domain gap in the Office-Home dataset poses a greater challenge in minimizing the second term of the upper bound in Equation~\ref{eq:theroemfromcourtyCatForoursDAL}. Note that CDAN achieves higher performance, likely because it employs class-conditional alignment techniques. By aligning the distributions based on the class information, CDAN is able to improve performance on the target domain.
Based on our observations, we can propose a potential direction for further improving our method in the future. By incorporating class-conditional alignment techniques, such as pseudo-labeling the target domain data, we may be able to enhance the performance of LDAuCID. This additional alignment strategy can provide more fine-grained alignment between the domains, resulting in improved performance, particularly on datasets with larger domain gaps like the Office-Home dataset.

  \begin{table*}[t!]
 \centering 
 \setlength{\tabcolsep}{1pt}
{\footnotesize
\begin{tabular}{lc|cccccccccccc|c}   
\multicolumn{2}{c}{Method}    &A$\rightarrow$C & A$\rightarrow$P & A$\rightarrow$R&C$\rightarrow$A&C$\rightarrow$P &C$\rightarrow$R &P$\rightarrow$A & P$\rightarrow$C & P$\rightarrow$R&R$\rightarrow$A&R$\rightarrow$C &R$\rightarrow$P & Average\\
\hline
\multicolumn{2}{c|}{Source Only~ \cite{he2016deep}}&  34.9&  50.0&  58.0&  37.4&  41.9&  46.2&  38.5&  31.2&  60.4&  53.9&  41.2&  59.9 & 46.1\\ 
\multicolumn{2}{c|}{DANN~ \cite{ganin2016domain}}&  45.6 &59.3& 70.1 &47.0& 58.5& 60.9 &46.1& 43.7& 68.5 &63.2 &51.8& 76.8   & 57.6\\ 
\multicolumn{2}{c|}{CDAN~ \cite{long2018conditional} }& 49.0& 69.3 &74.5 &\textbf{55.4}& 66.0& \textbf{68.4}& \textbf{55.6} &\textbf{48.3}& \textbf{75.9} &\textbf{68.4}& \textbf{55.4}& \textbf{80.5} &	 \textbf{63.9}\\ 
\multicolumn{2}{c|}{DAN~ \cite{long2015learning}}&  43.6& 57.0 &67.9& 45.8 &56.5& 60.4& 44.0& 43.6& 67.7& 63.1& 51.5 &74.3 & 56.3\\
\multicolumn{2}{c|}{JAN~ \cite{long2017deep}}&  45.9 &61.2& 68.9 &50.4& 59.7 &61.0& 45.8 &43.4& 70.3& 63.9& 52.4 &76.8&  58.3\\ 
\multicolumn{2}{c|}{DJT~ \cite{damodaran2018deepjdot}}& 39.7& 50.4& 62.4 &39.5& 54.3& 53.1 &36.7 &39.2& 63.5& 52.2& 45.4 &70.4&50.6 \\ 
\hline
\multicolumn{2}{c|}{LDAuCID }&	  \textbf{48.3} &\textbf{67.4}  & \textbf{74.1} & 48.7&  \textbf{61.9}&  63.8&   49.6&  42.1&  71.3&  60.3&  47.6& 76.6 & 59.4 \\
\end{tabular}}

\caption{ Classification accuracy for UDA tasks of  Office-Home dataset. }
\label{table:tabDA4}
 \end{table*}

Lastly, the performance comparison results for the Office-Caltech benchmark are presented in Table \ref{table:tabDA6}. It is worth noting the outstanding performance achieved by the LDAuCID algorithms on this benchmark. Across the majority of tasks within the benchmark, our approach demonstrates state-of-the-art performance. This   performance can be attributed to the relatively small domain gaps between the tasks in the Office-Caltech benchmark. The compact domain gaps imply that the upperbound  in Theorem 1, which constraints the performance of our approach, is   tighter. Consequently, the improved upperbound leads to enhanced performance and superior results in terms of accuracy and effectiveness.
We conclude that the performance for our algorithm is superior in regimes that the domain gap is smaller.
We conclude that due to the varying degrees of domain gaps across different datasets, tailored approaches might help to achieve optimal performance. The ability to adapt our algorithm to different benchmark requirements and domain gaps further solidifies its versatility and robustness.

  \begin{table*}[t!]
 \centering 
 \setlength{\tabcolsep}{1pt}
{\footnotesize
\begin{tabular}{lc|cccccccccccc|c}   
\multicolumn{2}{c}{Method}    &A$\rightarrow$C & A$\rightarrow$D & A$\rightarrow$W&W$\rightarrow$A&W$\rightarrow$D &W$\rightarrow$C &D$\rightarrow$A & D$\rightarrow$W & D$\rightarrow$C&C$\rightarrow$A&C$\rightarrow$W &C$\rightarrow$D& Average\\
\hline
\multicolumn{2}{c|}{Source Only}& 84.6 &81.1 &75.6 &79.8 &98.3 &79.6 &84.6 &96.8& 80.5 &92.4 &84.2& 87.7& 85.4 \\
\multicolumn{2}{c|}{DANN~ \cite{ganin2016domain}}&  87.8&  82.5&  77.8&  83.0&  \textbf{100} & 81.3&  84.7&  99.0 & 82.1 & 93.3 & 89.5&  91.2&  87.7   \\
\multicolumn{2}{c|}{MMAN~ \cite{ma2019deep}}& 88.7 & 97.5& \textbf{96.6}& 94.2& \textbf{100}& 89.4& \textbf{94.3}& 99.3& 87.9& 93.7& \textbf{98.3} & 98.1& 94.6 \\
\multicolumn{2}{c|}{RevGrad~ \cite{ganin2014unsupervised}}&85.7 &89.2& 90.8 &93.8& 98.7& 86.9& 90.6 &98.3 &83.7 &92.8& 88.1 &87.9& 88.9 \\
\multicolumn{2}{c|}{DAN~ \cite{long2015learning}}&84.1 &91.7& 91.8 &92.1 &\textbf{100}& 81.2 &90.0 &98.5 &80.3 &92.0 &90.6& 89.3& 90.1\\
\multicolumn{2}{c|}{CORAL~ \cite{sun2016return}}&  86.2 &91.2 &90.5 &88.4 &\textbf{100} &88.6 &85.8 &97.9& 85.4 &93.0 &92.6 &89.5& 90.8\\  
\multicolumn{2}{c|}{WDGRL~ \cite{shen2018wasserstein}}& 87.0& 93.7 &89.5 &93.7 &\textbf{100}& 89.4 &91.7 &97.9& 90.2 &93.5& 91.6 &94.7 &92.7 \\  
\hline
\multicolumn{2}{c|}{LDAuCID}&  \textbf{99.6} &\textbf{100.0} & 86.5 & \textbf{96.1}&  \textbf{100}& \textbf{99.8}& 88.5 & \textbf{100.0} &  \textbf{95.7} & \textbf{99.3} & 96.4  & \textbf{99.8} &  \textbf{96.8}
\end{tabular}}
\caption{ Performance comparison  for UDA tasks of  Office-Caltech dataset. }
\label{table:tabDA6}
 \end{table*}

After analyzing the results presented in Tables \ref{table:tabDA1} to \ref{table:tabDA6}, we can   assert that LDAuCID performs remarkably well across all UDA tasks, despite its focus on addressing the challenges of continual UDA where most of the source domain data points are inaccessible. While it is expected that LDAuCID's performance may be upperbounded by classic UDA algorithms, we have observed that it outperforms many of these UDA methods in some of the standard UDA tasks.
It is noteworthy that our primary motivation was to tackle UDA in a CL setting. However, based on our observations, we can conclude that LDAuCID is also well-suited for addressing classic UDA problems. The algorithm's   performance across various UDA tasks underscores its versatility and effectiveness.
Nevertheless, note that our evaluation was conducted within the scope of continual UDA learning. We believe that further advancements and the development of subsequent methods specifically tailored for this learning regime will enable a more comprehensive and detailed comparison among different approaches.

In summary, our findings demonstrate that LDAuCID competes favorably with classic UDA methods across multiple UDA tasks, despite its focus on addressing the challenges of continual UDA learning. This highlights the algorithm's potential for both CL and classic UDA scenarios. 

\subsubsection{Analytic and Ablative Studies}

In order to empirically study the impact of our algorithm, we conducted an analysis of the data point geometry in the embedding space. The geometry of the data points provides an approximation of the learned distributions in the output of the encoder, which can be considered as the empirical version of the internally learned distribution. To facilitate this analysis and enable 2D visualization, we utilized the UMAP  visualization tool~ \cite{mcinnes2018umap}.
By applying UMAP, we  reduce the dimensionality of the data representations in the embedding space. This reduction allowed us to visualize the data points in a 2D space and gain insights into their geometric relationships. 

\begin{figure}[tb!]
    \centering
           \begin{subfigure}[b]{0.22\textwidth}\includegraphics[width=\textwidth]{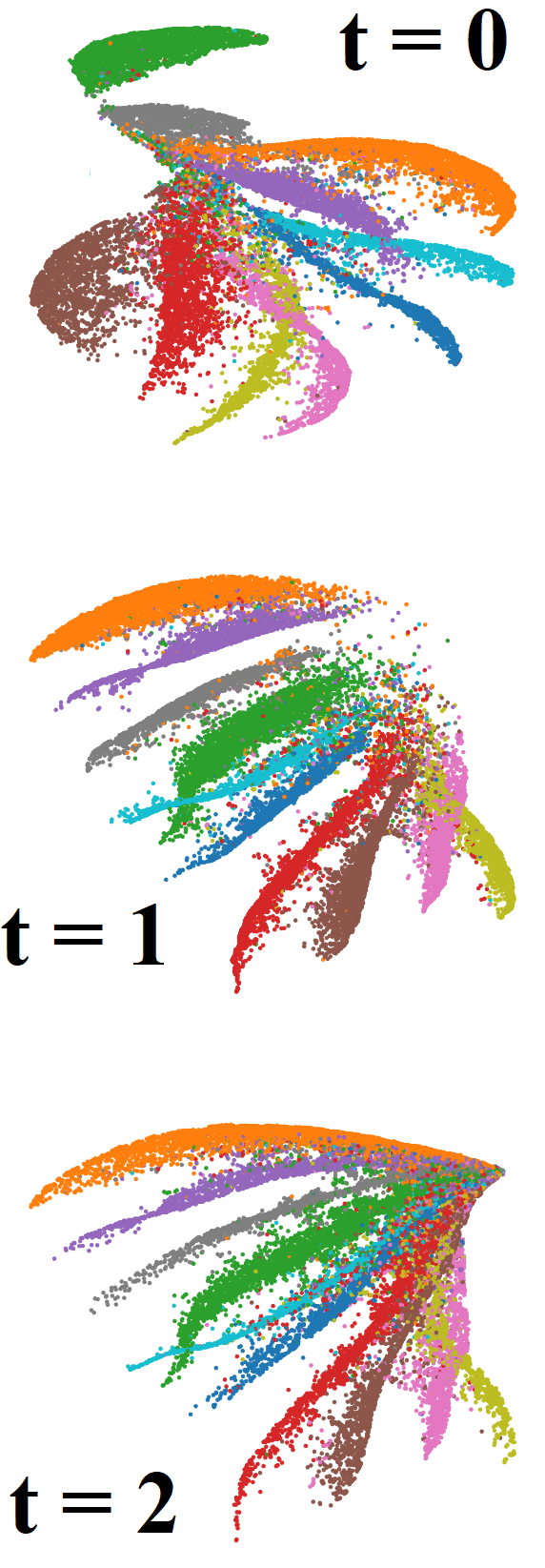}
           \centering
        \caption{SVHN}
        \label{NIPSDALfig:MNISTUSPS}
    \end{subfigure}
    \begin{subfigure}[b]{0.22\textwidth}\includegraphics[width=\textwidth]{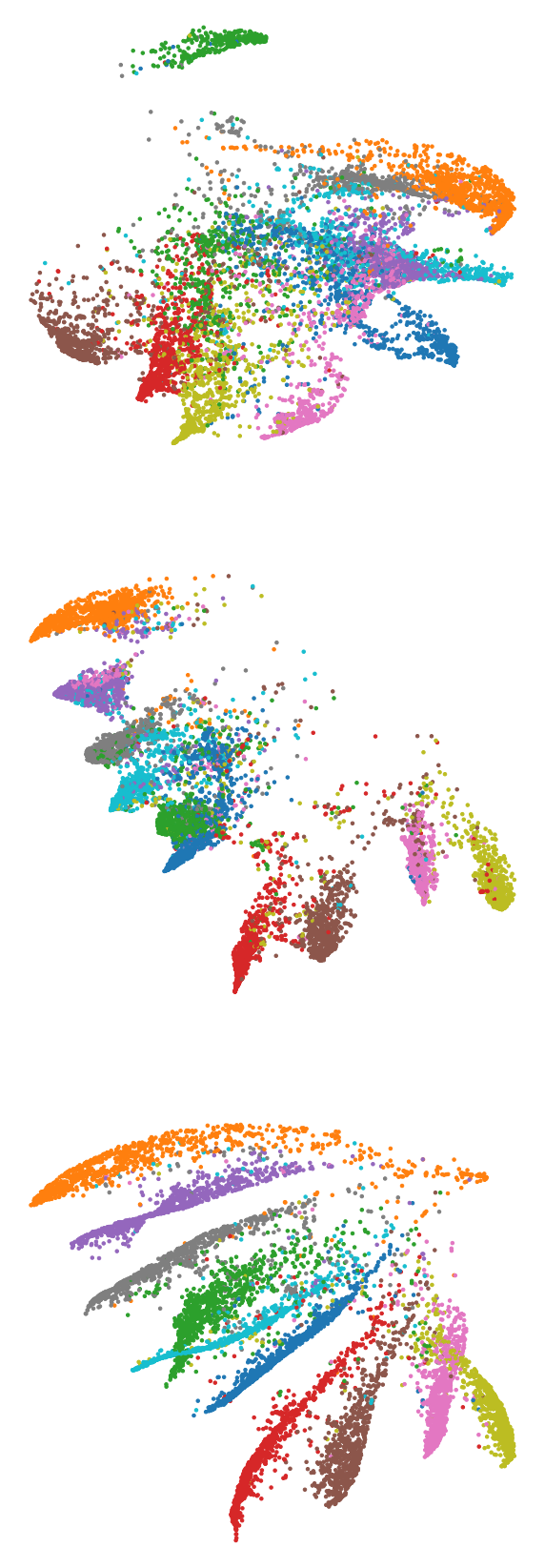}
           \centering
        \caption{MNIST}
        \label{NIPSDALfig:USPSMNIST}
    \end{subfigure}
       \begin{subfigure}[b]{0.22\textwidth}\includegraphics[width=\textwidth]{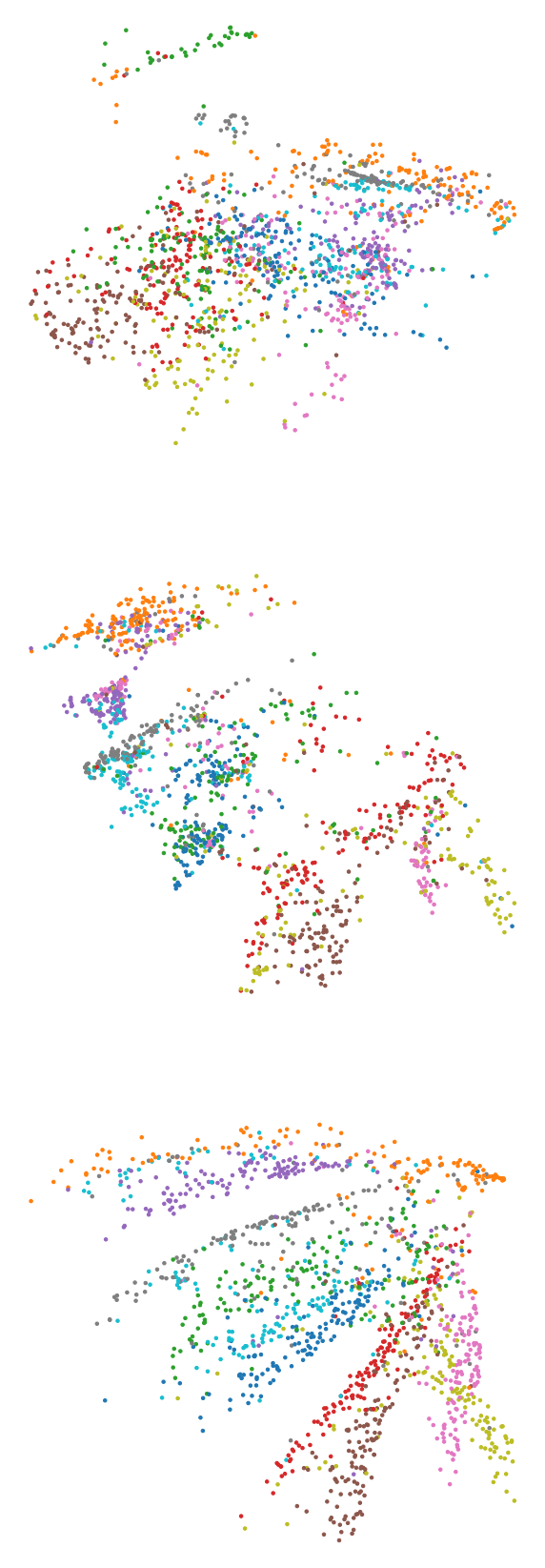}
           \centering
        \caption{USPS}
        \label{NIPSDALfig:MNISTUSPSembed}
    \end{subfigure}
       \begin{subfigure}[b]{0.22\textwidth}\includegraphics[width=\textwidth]{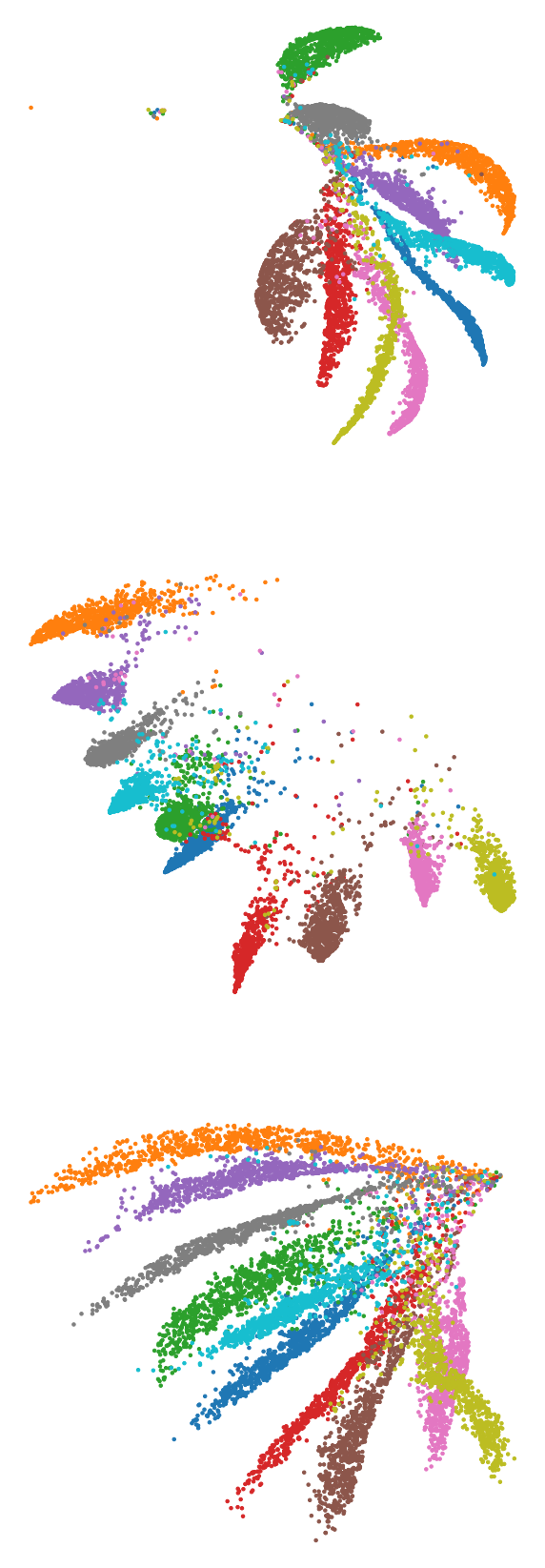}
              \centering
        \caption{GMM}
        \label{NIPSDALfig:USPSMNISTembed}
    \end{subfigure}
     \caption{(a-d) UMAP visualization for  the testing split of the domains in   the UDA task $\mathcal{S}\rightarrow \mathcal{M}\rightarrow  \mathcal{U}$   and the fitted GMM at   time-steps $t=0,1,2$. Visualizations at each of the rows are computed after learning the $t^{th}$ task prior to the time-step $t+1$.  ) }\label{NIPSDALfig:resultsCatforgetRelated1}
\end{figure}

Figure~\ref{NIPSDALfig:resultsCatforgetRelated1} presents a visualization of the testing splits of the source domain and two target domains, along with a selection of randomly drawn samples from the internally learned GMM. The purpose of this figure is to provide insights into the data geometry in the embedding space for the sequential digit recognition task from source domain ($\mathcal{S}$) to target domains ($\mathcal{M}$ and $\mathcal{U}$).
In the figure, each data point represents an individual data sample, and different colors are used to represent the ten different digit classes. Each row in Figure~\ref{NIPSDALfig:resultsCatforgetRelated1} corresponds to the data geometry at the end of a specific time-step ($t$), indicating the progress of learning over time. For example, the second row shows the data geometry after learning the SVHN and MNIST datasets.
By vertically examining the sub-figures within each column of the figure, we can observe the effect of learning additional tasks over time. Specifically, we can analyze how the learned knowledge is retained and how the separability of classes is preserved when the model is adapted to future tasks. This stability in data geometry suggests that our approach effectively mitigates catastrophic forgetting as the model is continuously updated.
In the last row of Figure~\ref{NIPSDALfig:resultsCatforgetRelated1}, we can see that all domains share a similar distribution, resembling the internally learned GMM distribution. This observation indicates that our method successfully aligns the distributions of different domains, making them share a common internal distribution.
As an example, if we compare the distribution of the MNIST dataset in the first row with the distribution in the second row, we can observe that, through domain adaptation, the distribution of MNIST becomes more similar to the SVHN dataset (source domain). This alignment of distributions in the embedding space validates our analytical deduction from Theorem~1.

 \begin{figure}[tb!]
    \centering
          \begin{subfigure}[b]{0.445\textwidth}\includegraphics[width=\textwidth]{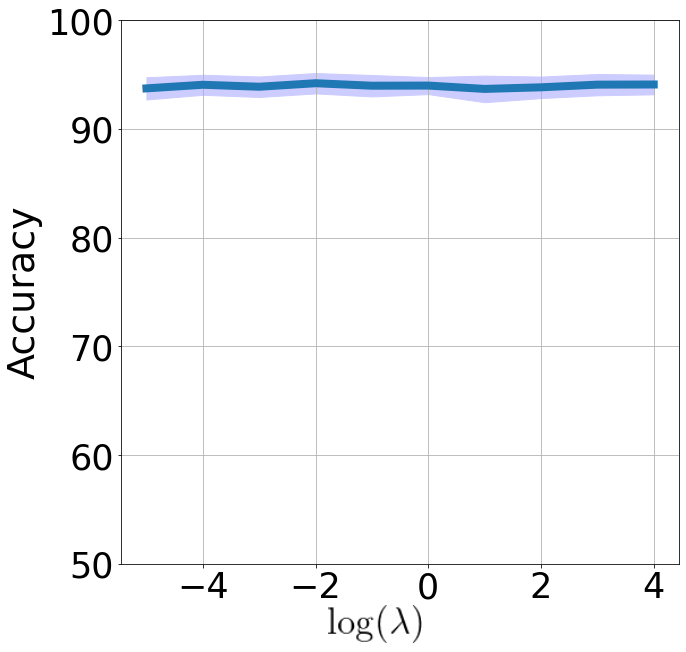}
           \centering
       \caption{ }
        \label{NIPSDALfig:CMafter}
    \end{subfigure}
      \centering
           \begin{subfigure}[b]{0.445\textwidth}\includegraphics[width=\textwidth]{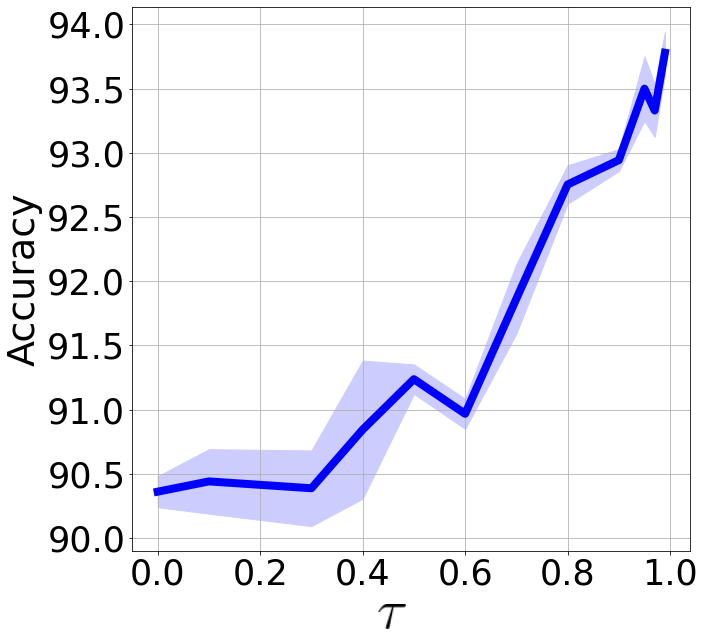}
           \centering
        \caption{ }
        \label{NIPSDALfig:CMtarget}
    \end{subfigure}
     \caption{  (e-f) Effect of the values for the hyperparameters $\lambda$ and $\tau$  on model generalization in the target domain for the binary UDA task $\mathcal{S}\rightarrow\mathcal{M}$. (Best viewed in color and on screen.) }\label{NIPSDALfig:resultsCatforgetRelated2}
\end{figure}


To achieve optimal performance, it is crucial to determine suitable values for the hyperparameters of our algorithm. We conducted experiments to examine the impact of the hyperparameters $\lambda$ and $\tau$ on the model's performance in the binary UDA task $\mathcal{S}\rightarrow\mathcal{M}$. Figures~\ref{NIPSDALfig:CMafter} and~\ref{NIPSDALfig:CMtarget} illustrate the results of these experiments, providing insights on how to tune these parameters for practical usage. In the figures, the dark curve represents the average performance, while the lighter shaded region around the curve represents the standard deviation.
In Figure~\ref{NIPSDALfig:CMafter}, we observe that the value of the parameter $\lambda$ has a minimal effect on the performance. This observation aligns with our expectations since the ERM loss term $\mathcal{L}(\cdot)$ in Eq. \eqref{eq:mainPrMatch} is relatively small due to pre-training on the source domain. Consequently, the primary optimization focus lies on the domain-alignment term in Eq. \eqref{eq:mainPrMatch}. Hence, precise fine-tuning of the trade-off parameter $\lambda$ is not essential for achieving satisfactory results.
On the other hand, Figure~\ref{NIPSDALfig:CMtarget} demonstrates that setting the confidence parameter $\tau$ to a value close to 1 leads to improved performance on the target domain. This observation aligns with our intuition as samples with high outlier scores in the GMM distribution can introduce label pollution, posing a challenge for domain alignment in UDA. Therefore, a higher confidence threshold helps mitigate the influence of potential outlier samples, resulting in better alignment performance.

We conducted additional controlled experiments on the $\mathcal{S}\rightarrow\mathcal{M}\rightarrow\mathcal{U}$ task, as presented in Table~\ref{table:tabDA5}, to gain better insights into the performance of our proposed algorithm.
In order to assess the robustness of our algorithm in the presence of data imbalance, we intentionally introduced imbalance across all domains in the $\mathcal{S}\rightarrow\mathcal{M}\rightarrow\mathcal{U}$ task. Specifically, we manipulated the dataset such that for each digit $i$, only $\frac{i+1}{10}$ portion of the data points were available. The results of these experiments are reported in Table~\ref{table:tabDA5}, where each row represents the performance of the algorithm at the end of the learning period for task $t$.
Despite the introduced data imbalance, we observed that our method still effectively mitigated catastrophic forgetting and yielded improved performance in the target domains. Although the overall performance was reduced compared to the balanced dataset scenario, our algorithm demonstrated its ability to adapt to the imbalanced dataset and achieve notable improvements in the target domains. These findings indicate that our method exhibits resilience and can handle data imbalance to a certain extent, reinforcing the practical applicability and robustness of LDAuCID.

Finally, we study the effect of the buffer size $N_b$ on the performance of the model. In our previous experiments, we arbitrarily set $N_b=10$ since, in comparison to the sizes of the datasets, this value is relatively small. However, it is important to investigate the impact of increasing the buffer size and understand how it influences the performance of our algorithm.
To study the effect of this hyperparameter, we conducted experiments by increasing the buffer size to $N_b=50$ and $N_b=100$. Performance results for this study are presented in Table~\ref{table:tabDA5}. As expected, we observed that using a larger buffer size led to improved performance and mitigated forgetting effects. This improvement can be attributed to the increased capacity of the buffer to store more informative samples, allowing the model to have better access to relevant data during the continual learning process. The result is also in line with our theoretical analysis. 
In practical applications, it is recommended to select the buffer size to be as large as possible, taking into account the hardware storage limitations. By increasing the buffer size, the model has access to a larger pool of past samples, which enhances its ability to mitigate catastrophic forgetting and maintain performance across tasks.

  \begin{table*}[t!]
 \setlength{\tabcolsep}{1pt}
 \centering 
{\small
\begin{tabular}{lc|ccc|c|ccc|c|ccc}   
\multicolumn{5}{c|}{Imbalanced Dataset} & \multicolumn{4}{c|}{$N_b=50$} & \multicolumn{4}{c}{$N_b=100$} \\
\hline
\multicolumn{2}{c|}{Time-step }    & $t=0$ & $t=1$ & $t=2$  & Time-step & $t=0$ & $t=1$ & $t=2$& Time-step & $t=0$ & $t=1$ & $t=2$\\
\hline
\multicolumn{2}{c|}{$\mathcal{S}$}&  89.8   &   84.8  & 80.2	  &$\mathcal{S}$&   92.3   & 83.6   & 82.0 &$\mathcal{S}$& 93.3   & 83.6   & 83.1  \\ 
\multicolumn{2}{c|}{$\mathcal{M}$}&  -    &  93.2   & 93.0   &$\mathcal{M}$&   -    & 94.2   & 95.6 & $\mathcal{M}$& -    & 95.6   & 97.1  \\ 
\multicolumn{2}{c|}{$\mathcal{U}$}&   -   &    -    & 93.1	  &$\mathcal{U}$&  - &-  & 91.9 &$\mathcal{U}$& - & - &94.6\\ 
\end{tabular}}
\caption{Analytic experiments using the $\mathcal{S}\rightarrow\mathcal{M}\rightarrow\mathcal{U}$ digit recognition task.    }
\label{table:tabDA5}
 \end{table*}

 \section{Conclusions  }
 We propose an algorithm for domain adaptation in the context of continual learning. Our approach is based on the assumption that a neural classifier maps the input distribution to an internal distribution in an embedding space, represented by a hidden layer of a neural network. This internal distribution is multimodal, where each mode represents one of the classes. The main objective of our method is to consolidate this internally learned distribution, ensuring that all learned tasks share a similar distribution in the embedding space. By achieving this objective, our algorithm preserves the model's generalizability when new tasks are learned, and effectively mitigates catastrophic forgetting.

To address continual domain adaptation, we employ experience replay as a means to alleviate catastrophic forgetting. We store informative input samples that contribute to estimating the internally learned distribution and replay them during subsequent training phases. We select these samples to be representative of the internal distribution. This selection allows the model to retain important knowledge from past tasks, enhancing its performance when adapting to new domains. We utilize a straightforward approach for estimating the internal distribution. However, we acknowledge that more sophisticated techniques for estimating the internal distribution could potentially improve the performance of continual learning, particularly when the internal distribution modes cannot be approximated well with a single Gaussian.
In terms of future research, we identify two potential areas of exploration. Firstly, we can investigate the optimal task order for the best performance in continual learning. Understanding how the sequential order of tasks affects the model's ability to adapt and retain knowledge will provide valuable insights for optimizing the learning process. Ideally, a CL algorithm should be designed such that any task order can be handled. Secondly,   extending our algorithm to the incremental learning setting, where new classes can be introduced beyond the initial training phase, can be another research direction. This extension would allow our method to handle scenarios where the model needs to adapt to novel classes over time.
 
 \section*{CRediT authorship contribution statement}
Mohammad Rostami contributed to  conceptualization, methodology, validation, writing – original draft, and writing – review \& editing.

 \section*{Declaration of competing interest}
The author declares that he does not have  known competing financial interests   that could impacted the results reported in this work.

 {
    \small
    \bibliographystyle{plain}
    \bibliography{ref}
}

\clearpage
\appendix

 \section{Sliced Wasserstein distance}

We have employed the Sliced Wasserstein (SWD) distance to assess distribution dissimilarity. For those curious, we offer a brief overview here. SWD is formulated on the basis of the Wasserstein distance (WD). The Wasserstein distance between two probability distributions $p_{\mathcal{S}}$ and $p_{\mathcal{T}}$ can be articulated as follows:
\begin{equation}
W_c(p_{\mathcal{S}},p_{\mathcal{T}})=\text{inf}_{\gamma\in \Gamma(p_{\mathcal{S}},p_{\mathcal{T}})} \int_{{X}\times {Y}} c(x,y)d\gamma(x,y).
\label{eq:kantorovich}
\end{equation}
Here, $\Gamma(p_{\mathcal{S}}, p_{\mathcal{T}})$ represents the set of all joint distributions $p_{{\mathcal{S}},{\mathcal{T}}}$ with marginal single-variable distributions $p_{\mathcal{S}}$ and $p_{\mathcal{T}}$. The function $c:X\times Y\rightarrow \mathbb{R}^+$ denotes the transportation cost, typically assumed to be the Euclidean distance in $\ell_2$-norm.

The computation of Wasserstein distance   entails solving a linear programming problem because both the objective function in Eq. ~\eqref{eq:kantorovich} and the constraint on $\gamma$ are linear. Specifically, when the distributions are one-dimensional, the computation of WD simplifies to a closed-form solution, as outlined below:
\begin{equation}
W_c(p_{\mathcal{S}},p_{\mathcal{T}})= \int_{0}^1 c(P_{\mathcal{S}}^{-1}(\tau),P_{\mathcal{T}}^{-1}(\tau))d\tau.
\label{eq:oneD}
\end{equation} 
Here, $P_{\mathcal{S}}$ and $P_{\mathcal{T}}$ denote the cumulative distributions of the one-dimensional distributions $p_{\mathcal{S}}$ and $p_{\mathcal{T}}$. The derived closed-form solution, characterized by significantly lower computational complexity in contrast to Eq. \eqref{eq:kantorovich}, serves as a driving force behind introducing the Sliced Wasserstein Distance (SWD). This motivation stems from the aim to expand the utility of the one-dimensional Eq. \eqref{eq:oneD} to distributions of higher dimensions. This broadened scope enhances the applicability of SWD across various scenarios.

The concept underlying SWD draws inspiration from slice sampling~ \cite{neal2003slice}. The approach involves projecting two $d$-dimensional probability distributions onto their respective marginal one-dimensional distributions, essentially slicing the high-dimensional distributions. The Wasserstein distance is then approximated by integrating the Wasserstein distances between the resulting one-dimensional marginal probability distributions across all possible one-dimensional subspaces, each of which possesses a closed-form solution. This approximation proves effective for optimal transport, given that any probability distribution can be uniquely represented through the set of one-dimensional marginal projection distributions~ \cite{helgason2011radon}.
For the distribution $p_\mathcal{S}$, a one-dimensional slice of the distribution is defined as follows:
\begin{equation}
\mathcal{R}p_\mathcal{S}(t;\bm{\gamma})=\int_{\mathcal{S}^{d-1}} p_\mathcal{S}(\bm{x})\bm{\delta}(t-\langle\bm{\gamma}, \bm{x}\rangle)d\bm{x},
\label{eq:radon}
\end{equation}
Here, $\bm{\delta}(\cdot)$ represents the Kronecker delta function, $\langle \cdot ,\cdot\rangle$ denotes the dot product of vectors, $\mathbb{S}^{d-1}$ is the $d$-dimensional unit sphere, and $\bm{\gamma}$ is the projection direction. Specifically, $\mathcal{R}p_\mathcal{S}(\cdot;\bm{\gamma)}$ denotes a marginal distribution of $p_\mathcal{S}$ obtained by integrating $p_\mathcal{S}$ over hyperplanes orthogonal to $\bm{\gamma}$. To put it differently, it represents the projection of $p_\mathcal{S}$ onto the hyperplanes determined by $\bm{\gamma}$. SWD is then defined as the integral of the Wasserstein distance between these sliced distributions, considering all possible one-dimensional subspaces $\bm{\gamma}$ on the unit sphere:
\begin{eqnarray}
SW(p_\mathcal{S},p_\mathcal{T})=   \int_{\mathbb{S}^{d-1}} W(\mathcal{R} p_\mathcal{S}(\cdot;\gamma),\mathcal{R} p_\mathcal{T}(\cdot;\gamma))d\gamma.
\label{eq:radonSWDdistance}
\end{eqnarray}
Here, $W(\cdot)$ represents the Wasserstein distance. The primary advantage of employing SWD, as evident from Eq~\eqref{eq:radonSWDdistance}, lies in the fact that, unlike the Wasserstein distance, SWD computation doesn't necessitate a computationally intensive optimization process. This aspect is attributed to the closed-form solution for the Wasserstein distance between two one-dimensional probability distributions. 

Since only samples from distributions are available, the one-dimensional Wasserstein distance can be approximated by the $\ell_p$-distance between the sorted samples.
This approach allows us to compute the integrand function in Eq. \eqref{eq:radonSWDdistance} for a given $\gamma$. To approximate the integral in Eq. \eqref{eq:radonSWDdistance}, a Monte Carlo-style integration can be employed. Firstly, we sample the projection subspace $\bm{\gamma}$ from a uniform distribution defined over the unit sphere and then compute the one-dimensional WD on the sample. The integral in Eq. ~\eqref{eq:radonSWDdistance} can then be approximated by computing the arithmetic average over a sufficiently large number of drawn samples.
In a formal sense, the SWD between $f$-dimensional samples $\{\phi(\bm{x}i^\mathcal{S})\in \mathbb{R}^f\sim p_\mathcal{S}\}_{i=1}^M$ and $\{\phi(\bm{x}i^\mathcal{T})\in \mathbb{R}^f \sim p_\mathcal{T}\}_{j=1}^M$  can be approximated as the following:
\begin{equation}
SW^2(p_\mathcal{S},p_\mathcal{T})\approx \frac{1}{L}\sum_{l=1}^L \sum_{i=1}^M| \langle\gamma_l, \phi(\bm{x}_{s_l[i]}^\mathcal{S}\rangle)- \langle\gamma_l, \phi(\bm{x}_{t_l[i]}^\mathcal{T})\rangle|^2.
\label{eq:SWDempirical}
\end{equation}
Here, $\gamma_l\in\mathbb{S}^{f-1}$ represents a uniformly drawn random sample from the unit $f$-dimensional ball $\mathbb{S}^{f-1}$, while $s_l[i]$ and $t_l[i]$ denote the sorted indices of $\{\gamma_l\cdot\phi(\bm{x}i)\}_{i=1}^M$ for the source and target domains, respectively. We employ the empirical form of SWD as expressed in Eq. \eqref{eq:SWDempirical} as the measure of dissimilarity between the probability distributions to align them in the embedding space. Notably, the function in Eq. \eqref{eq:SWDempirical} is differentiable concerning the encoder parameters. Consequently, we can leverage gradient-based optimization techniques commonly utilized in deep learning to minimize it with respect to the model parameters.

\end{document}